%% file: main_jpdc.tex
\documentclass[a4paper,fleqn]{cas-dc}

\usepackage[numbers]{natbib}

\usepackage{amsmath,amssymb,amsthm}
\usepackage{algorithm}
\usepackage{algpseudocode}
\usepackage{booktabs}
\usepackage{multirow}
\usepackage[caption=false]{subfig}
\usepackage{array}
\usepackage{stfloats} % allow full-width floats at column bottoms
\usepackage{placeins} % one \FloatBarrier before Discussion bounds any residual float drift
\usepackage{capt-of} % \captionof for non-floating inline figures (minimal; cas-dc-safe)
\usepackage{framed}  % leftbar callout for set-off key statements
\usepackage{tikz}
\usetikzlibrary{arrows.meta,positioning,calc,backgrounds,fit,shapes.geometric}
\newcommand{\clientnode}[2]{\node[circle,draw=orange!80!black,fill=orange!75,thick,
   minimum size=1.0cm,font=\footnotesize] (#1) at #2 {Client};}
\newcommand{\honnode}[2]{\node[circle,draw=orange!80!black,fill=orange!70,minimum size=0.42cm] (#1) at #2 {};}
\newcommand{\byznode}[2]{%
  \node[circle,draw=red!70!black,fill=red!80,minimum size=0.44cm] (#1) at #2 {};
  \fill[red!80] ($(#1)+(-0.13,0.1)$)--($(#1)+(-0.06,0.3)$)--($(#1)+(0.0,0.12)$)--cycle;
  \fill[red!80] ($(#1)+(0.13,0.1)$)--($(#1)+(0.06,0.3)$)--($(#1)+(0.0,0.12)$)--cycle;}
\newcommand{\modelsq}[1]{\begin{scope}[shift={#1}]
  \fill[blue!22](-0.22,-0.22)rectangle(0.22,0.22);
  \draw[blue!55](-0.22,-0.073)--(0.22,-0.073)(-0.22,0.073)--(0.22,0.073)
                (-0.073,-0.22)--(-0.073,0.22)(0.073,-0.22)--(0.073,0.22);
  \draw[blue!70,thick](-0.22,-0.22)rectangle(0.22,0.22);\end{scope}}
\newcommand{\sketchsq}[2]{\begin{scope}[shift={#1}]
  \fill[green!50!black!16](-0.15,-0.15)rectangle(0.15,0.15);
  \draw[green!55!black!45](-0.15,0)--(0.15,0)(0,-0.15)--(0,0.15);
  \draw[#2,thick](-0.15,-0.15)rectangle(0.15,0.15);\end{scope}}
\newcommand{\okmark}[1]{\draw[green!55!black,line width=1.1pt]
  ($#1+(-0.08,0)$)--($#1+(-0.02,-0.08)$)--($#1+(0.1,0.09)$);}
\newcommand{\xmark}[1]{\draw[red!75,line width=1.1pt] ($#1+(-0.07,0.07)$)--($#1+(0.07,-0.07)$);
  \draw[red!75,line width=1.1pt] ($#1+(-0.07,-0.07)$)--($#1+(0.07,0.07)$);}

\newtheorem{theorem}{Theorem}
\newtheorem{lemma}{Lemma}

\newtheorem{assumption}{Assumption}
\theoremstyle{definition}
\newtheorem{definition}{Definition}

\newcommand{\R}{\mathbb{R}}
\newcommand{\N}{\mathcal{N}}
\newcommand{\CS}{\mathrm{CS}}
\newcommand{\Sacc}{\mathcal{S}}
\newcommand{\geff}{\gamma_{\mathrm{eff}}}
\newcommand{\bw}{\mathbf{w}}
\newcommand{\bs}{\mathbf{s}}
\newcommand{\bv}{\mathbf{v}}
\newcommand{\bu}{\mathbf{u}}

\makeatletter
\ExplSyntaxOn
\NewDocumentEnvironment{inlinefigure}{}
  {
    \par\medskip
    \__reset_fig:
    \dim_set:Nn \l_fig_width_dim { \linewidth }
    \csgdef{figwidth}{\dim_use:N \l_fig_width_dim}
    \cs_set_eq:NN \@makecaption \__make_fig_caption:nn
    \def\@captype{figure}
    \centering\sffamily\small
  }
  { \par\medskip }
\ExplSyntaxOff
\makeatother
\newcommand{\gin}{\gamma_{\mathrm{in}}}
\input{sections/rev_macros}
\input{sections/rev_macros_cont}
\input{sections/rev_macros_scale}
\begin{document}

\let\WriteBookmarks\relax
\def\floatpagepagefraction{1}
\def\textpagefraction{.001}

\shorttitle{Screen before you fetch: compressed Byzantine screening}
\shortauthors{M.\ Rangwala et al.}

\title[mode=title]{Screen Before You Fetch: Compressed Byzantine Screening for Decentralized Learning on the Edge--Cloud Continuum}

\input{sections_jpdc/authors}

\input{sections_jpdc/abstract}


\maketitle

\input{sections_jpdc/intro}
\input{sections_jpdc/background}
\input{sections_jpdc/design}
\input{sections_jpdc/convergence}
\input{sections_jpdc/evaluation}
\input{sections_jpdc/discussion}
\input{sections_jpdc/conclusion}

\printcredits

\bibliographystyle{model1-num-names}
\bibliography{refs}

\input{sections_jpdc/appendix}
\end{document}

%% file: sections/rev_macros.tex
\newcommand{\decAgreeHon}{100.00}
\newcommand{\decLost}{0.000}
\newcommand{\distStd}{0.092}

\newcommand{\ampctsexOneZero}{11}

\newcommand{\ampctsexTwoZero}{21}

\newcommand{\ampctsexThreeZero}{31}

\newcommand{\ampctsexFourZero}{41}

\newcommand{\qBALqEightthirty}{0.25}
\newcommand{\qCtSexzero}{1.00}
\newcommand{\qCtSexthirty}{0.70}

\newcommand{\qCtSexqEightthirty}{0.17}
\newcommand{\resbandRone}{17.6}
\newcommand{\resBytesbandRone}{1299}
\newcommand{\resbandRfive}{3.4}
\newcommand{\resBytesbandRfive}{1228}
\newcommand{\resselectaftercommitRone}{66.4}
\newcommand{\resBytesselectaftercommitRone}{1557}
\newcommand{\resselectaftercommitRfive}{33.8}
\newcommand{\resBytesselectaftercommitRfive}{1389}

\newcommand{\sccCompressGap}{0.002}
\newcommand{\sccCompressN}{10}
\newcommand{\sccBytesMin}{0.70}
\newcommand{\scBigSpikeTER}{0.299}
\newcommand{\scLongRawTER}{0.320}

\newcommand{\scLongExTER}{0.335}

\newcommand{\scPassedRfive}{0}
\newcommand{\scPassedRaw}{20}
\newcommand{\scAttempts}{16660}

\newcommand{\llmAmpRatio}{1.70}

\newcommand{\gennullspacetopktopkbalancerawTER}{0.940}

\newcommand{\gennullspacetopktopkbalanceexTER}{0.292}

\newcommand{\genbandtopkctsrawTER}{0.858}
\newcommand{\genbandtopkctsrawPass}{53.4}
\newcommand{\genbandtopkctsexTER}{0.293}

\newcommand{\memnistgaussianBytes}{0.70}

\newcommand{\memnistalieBytes}{1.00}

\newcommand{\memnistbandRaw}{0.682}
\newcommand{\memnistbandEx}{0.292}

\newcommand{\mfemnistbandBal}{0.499}
\newcommand{\mfemnistbandRaw}{0.893}
\newcommand{\mfemnistbandEx}{0.499}

\newcommand{\mcifarOneZerobandBal}{0.482}
\newcommand{\mcifarOneZerobandRaw}{0.591}
\newcommand{\mcifarOneZerobandEx}{0.485}

\newcommand{\parityMaxGap}{0.003}

\newcommand{\ampfemnistPubRawTER}{0.935}
\newcommand{\ampfemnistPubExTER}{0.492}
\newcommand{\ampfemnistPubExBytes}{1.00}
\newcommand{\forgedfemnistNoBanBytes}{1.00}
\newcommand{\forgedfemnistBanBytes}{0.70}
\newcommand{\decfemnistAgreeHon}{100.00}
\newcommand{\ampcifarOneZeroPubRawTER}{0.686}
\newcommand{\ampcifarOneZeroPubExTER}{0.483}
\newcommand{\ampcifarOneZeroPubExBytes}{1.00}
\newcommand{\forgedcifarOneZeroNoBanBytes}{1.00}
\newcommand{\forgedcifarOneZeroBanBytes}{0.70}
\newcommand{\deccifarOneZeroAgreeHon}{100.00}
\newcommand{\topoSparseBal}{0.303}
\newcommand{\topoDenseBal}{0.282}
\newcommand{\hetNonIIDBal}{0.489}
\newcommand{\hetIIDBal}{0.260}
\newcommand{\topoSparseSg}{0.301}
\newcommand{\topoDenseSg}{0.282}
\newcommand{\hetNonIIDSg}{0.491}
\newcommand{\hetIIDSg}{0.259}

\newcommand{\txtQeightCtsDiff}{0.003}
\newcommand{\txtRfiveDiff}{0.001}
\newcommand{\txtBandRawTER}{0.682\pm0.088}
\newcommand{\txtBandBalTER}{0.292\pm0.006}
\newcommand{\txtBandExTER}{0.292\pm0.004}
\newcommand{\txtBandPass}{17.6}
\newcommand{\txtBandEightTER}{0.91}
\newcommand{\txtBandNineFiveBal}{0.901\pm0.008}
\newcommand{\txtBandNineFiveEx}{0.859\pm0.018}
\newcommand{\txtBandNineFiveRej}{26}
\newcommand{\txtScRate}{0.12}
\newcommand{\txtSacOne}{17.4}
\newcommand{\txtSacSixtyFour}{84.9}
\newcommand{\txtSacTERlo}{0.291}
\newcommand{\txtSacTERhi}{0.293}
\newcommand{\txtAmpPubRawFourZero}{0.979\pm0.002}
\newcommand{\txtLlmNullTER}{0.234\pm0.005}
\newcommand{\txtAgreeMin}{99.88}
\newcommand{\txtLongBal}{0.228\pm0.004}
\newcommand{\txtLongCts}{0.234\pm0.008}
\newcommand{\txtGammaTwoBand}{0.90}
\newcommand{\txtGammaOneBand}{0.57}
\newcommand{\txtGammaHalfBand}{0.34}

\newcommand{\forgedCtSTER}{0.292}
\newcommand{\forgedCtSMismatch}{5553}

\newcommand{\forgedCtSbanMismatch}{159}
\newcommand{\forgedCtSbanHonMismatch}{0}
\newcommand{\forgedCtSbanBans}{159}

\newcommand{\forgedCtSRatio}{1.00}

\newcommand{\forgedCtSbanRatio}{0.70}

%% file: sections/rev_macros_cont.tex
\newcommand{\contHubRelbalancerelayT}{0.19}
\newcommand{\contHubRelctsT}{0.82}
\newcommand{\contHubRelctsrelayT}{0.14}
\newcommand{\contBeaconTwoZeroDrop}{23}
\newcommand{\contBeaconThreeZeroDrop}{0}
\newcommand{\contBeaconFourFiveDrop}{0}

\newcommand{\contRelctsbZeroT}{1.21}

\newcommand{\contRelctsrelaybZeroT}{0.98}

\newcommand{\contRelbalancerelaybThreeZeroT}{0.95}

\newcommand{\contRelpublicbThreeZeroT}{1.01}

\newcommand{\contRelctsbThreeZeroT}{0.74}
\newcommand{\contRelctsbThreeZeroB}{0.68}
\newcommand{\contRelctsrelaybThreeZeroT}{0.61}

\newcommand{\contUpTwoBalT}{138}
\newcommand{\contUpTwoCtsT}{72}

%% file: sections/rev_macros_scale.tex
\newcommand{\scaleComputeMin}{3.6}
\newcommand{\scaleComputeMax}{5.1}
\newcommand{\scaleBEMin}{3.9}
\newcommand{\scaleBEMax}{7.4}
\newcommand{\scaleBytesRatio}{0.50}
\newcommand{\scaleBigBalT}{25}
\newcommand{\scaleBigCtsT}{89}
\newcommand{\scaleBigBalB}{72}
\newcommand{\scaleBigCtsB}{36}
\newcommand{\scaleBigBalX}{191}
\newcommand{\scaleBigCtsX}{95}

\newcommand{\scaleBigCtsShare}{1.5}
\newcommand{\scaleShareMax}{1.7}

%% file: sections_jpdc/authors.tex
\author[1]{Murtaza Rangwala}[orcid=0009-0003-4578-8671]
\cormark[1]
\ead{mrangwala@student.unimelb.edu.au}
\credit{Conceptualization, Methodology, Software, Formal analysis, Investigation, Writing -- original draft}
\affiliation[1]{organization={Quantum Cloud Computing and Distributed Systems (qCLOUDS) Lab, School of Computing and Information Systems, The University of Melbourne},
    city={Melbourne},
    country={Australia}}

\author[1,2]{Farag Azzedin}
\credit{Methodology, Writing -- review and editing}
\affiliation[2]{organization={Department of Information and Computer Science, King Fahd University of Petroleum and Minerals},
    city={Dhahran},
    country={Saudi Arabia}}

\author[1]{Richard O. Sinnott}
\credit{Supervision, Writing -- review and editing}

\author[1]{Rajkumar Buyya}
\credit{Supervision, Methodology, Writing -- review and editing}

\cortext[cor1]{Corresponding author}

%% file: sections_jpdc/abstract.tex
\begin{abstract}
In decentralized federated learning, Byzantine-robust defenses filter neighbors by model similarity,
but only after receiving each neighbor's full model. On constrained edge uplinks, that transfer
dominates the round, even for discarded models. We propose SketchGuard, which screens
neighbors on compact sketches and fetches full models only from those that pass, so a rejected
neighbor costs $O(k)$ rather than $O(d)$ where $k\ll d$. Sketches, however, hide information an adversary can
exploit, either by building a model indistinguishable from a benign one under a public compressor
or by forging a sketch once the seed is known. Three rules close these gaps,
namely exact verification of every fetched model, commitment before the sketch seed is drawn, and a
ban on any sender whose model mismatches its sketch. We prove convergence in the strongly convex
and non-convex settings with explicit topology and heterogeneity terms. Across three datasets and two
wrapped filters, SketchGuard reproduces the uncompressed filter's test error to within
\parityMaxGap, withstands a band-targeted attack that defeats sketch-only screening, and moves
$\qCtSexthirty\times$ the bytes of full exchange under attack for under \scaleShareMax\% of a round
in compute. On emulated edge and cloud links it cuts the round to $\contRelctsrelaybThreeZeroT\times$
that of the state-of-the-art.
\end{abstract}

%% file: sections_jpdc/intro.tex
\section{Introduction}
\label{sec:intro}

Federated learning (FL) trains a shared model across many data holders while each holder keeps its
raw data local~\cite{mcmahan2017}. The canonical architecture, however, relies on a server to
aggregate the clients' updates, which makes that server a single point of failure, a communication
bottleneck, and a party every client must trust to behave honestly~\cite{fang2024}. Decentralized FL
(DFL) removes the server. Instead, clients exchange model updates directly with their neighbors over
a communication graph, so no central authority exists and no single failure stops
training~\cite{lian2017,martinez2023}. A power utility, for instance, can train a fault-detection
model this way across its substations, with sensor data never leaving the site that recorded
it~\cite{li2022smartgrid}. The machines that take part in such a deployment are rarely alike. A few
are cloud instances in the utility's data center on gigabit links, whereas most are substation
gateways on metered cellular uplinks measured in single-digit megabits, the heterogeneous setting of
the edge--cloud continuum~\cite{rosendo2022,lim2020}.

Removing the server also introduces a distinctive adversarial challenge, achieving Byzantine
robustness. With no
gatekeeper, each client must decide for itself which neighbors' updates to trust, and a malicious
neighbor can send arbitrary or carefully crafted models to poison training or to drag the honest
clients away from consensus~\cite{krum2017,baruch2019,zhang2026smartgrid}. Moreover, each client
sees only its own neighborhood, often over non-IID data and changing connections. Consequently, the
global robust aggregation rules developed for server-based learning do not apply.

In DFL, most defenses rely on similarity-based filtering, in which a client accepts neighbor $j$'s
model only if it is sufficiently close to the client's
own~\cite{guo2022,he2022,fang2024,cajaraville2025}. The approach is well founded, and methods such as
BALANCE~\cite{fang2024} and SCCLIP~\cite{he2022} provide convergence guarantees. Nevertheless, it
rests on a costly architectural assumption. Every neighbor's full model must be received and
compared before any filtering decision is made, so a client with $|\N_i|$ neighbors receives
$O(d|\N_i|)$ parameters per round, however many of those models it then rejects.

This cost also grows linearly with node degree and with model size, regardless of the attack. A
node with 299 neighbors and a 60M-parameter model, for example, receives $\scaleBigBalB$\,GB per
round before it makes a single filtering decision. The continuum makes this worse, because the cost
falls on the weakest links. A gigabit cloud peer absorbs it, whereas a gateway on a megabit-class
link spends most of the round receiving models it might eventually reject.

One way to avoid this cost is to compare compact summaries instead of full models. A Count Sketch
(CS)~\cite{charikar2002} compresses a model of $d$ parameters into $k\ll d$ numbers through a random
mapping, and the distance between two sketches approximates the distance between the models. Since
filtering compares distances, a decision taken on sketches should mirror the decision taken on full
models. The expensive exchange can therefore wait until after filtering and happen only for accepted
neighbors.

However, the summary then becomes something the adversary can exploit. Because a sketch is far
smaller than the model, many different models share the same sketch, and an adversary who knows how
sketches are computed can build a harmful model whose sketch looks benign. The consequence depends
on whether the defender checks only the sketch or also the full model it downloads. A client that
trusts the sketch aggregates the harmful model and is poisoned. Conversely, a client that rechecks
every downloaded model stays safe, but it now downloads every malicious neighbor in full, every
round, and the saving that motivated compression is gone. Hiding the sketch's random seed until
models are fixed does not end the matter either, because each sender reports its own sketch and can
simply lie about it. In each case, the adversary acts after it has learned something the defender
relies on.

To close these gaps, we propose SketchGuard, which addresses each of them with one rule. Every
downloaded model is rechecked by the original filter, so the sketch decides only what is
downloaded, never what is trusted. Every client commits to its model before the round's sketch seed
is drawn, so no model can be built against the sketch. A sender whose model does not match the
sketch it reported is banned. SketchGuard wraps any similarity-based filter with a Euclidean
acceptance radius~\cite{fang2024,he2022,guo2022,pillutla2022}, and we instantiate it with the
state-of-the-art BALANCE~\cite{fang2024} for analysis and evaluation. Our main contributions are the
following.

\begin{itemize}
\item \textbf{Algorithm.} SketchGuard reduces per-round communication from $O(d|\N_i|)$ to
$O(k|\N_i|+d|\Sacc_i|)$ by exchanging sketches for screening and full models only for accepted
neighbors (Section~\ref{sec:protocol}).
\item \textbf{Security.} We show that, for any public linear compressor, compressed screening is
either poisoned or forced to download every Byzantine neighbor in full
(Theorem~\ref{thm:public}), and that the three rules close these failures
(Lemma~\ref{lem:exact}, Theorems~\ref{thm:cts} and~\ref{thm:ban}).
\item \textbf{Theory.} We prove convergence in the strongly convex and non-convex settings, with a
rate that exposes the spectral gap and data heterogeneity (Theorems~\ref{thm:onestep}
to~\ref{thm:topo}). The compressor does not appear in the bounds.
\item \textbf{Experiments.} On three image datasets, a language model, and two wrapped filters,
SketchGuard matches the accuracy of the uncompressed filter under every attack we test, including
attacks designed against the protocol itself. Under attack it moves $\qCtSexthirty\times$ the bytes
of full exchange, and on emulated edge and cloud links it cuts the round time to
$\contRelctsrelaybThreeZeroT\times$ that of BALANCE.
\end{itemize}

\noindent The rest of the paper is organized as follows. Section~\ref{sec:background} gives the
preliminaries and related work. Section~\ref{sec:design} presents SketchGuard, its security, and its
cost. Section~\ref{sec:convergence} analyzes convergence, Section~\ref{sec:eval} reports the
evaluation, Section~\ref{sec:discussion} discusses scope, adaptive attacks, deployment, and
limitations, and Section~\ref{sec:conclusion} concludes with future directions.

%% file: sections_jpdc/background.tex
\section{Preliminaries and related work}
\label{sec:background}

\subsection{DFL problem formulation and protocol}
\label{sec:formulation}

Consider $n$ clients connected by an undirected graph $G=(V,E)$. The clients span
bandwidth-constrained edge devices and well-provisioned cloud peers, on links whose capacities
differ by orders of magnitude. Client $i$ holds a private dataset $\mathcal D_i$ and a local model
$\bw_i\in\R^d$. We write $H$ for the honest clients and $B$ for the Byzantine ones. The honest
clients jointly minimize
\begin{equation}
\min_{\bw\in\R^d} F(\bw)=\frac1{|H|}\sum_{i\in H} f_i(\bw),
\end{equation}
where $f_i(\bw)=\mathbb E_{(\mathbf x,y)\sim\mathcal D_i}[\ell(\bw;\mathbf x,y)]$ is the expected loss
on client $i$'s data. Each round alternates two steps. Every client first takes a local stochastic
gradient step,
\begin{equation}\label{eq:local-update}
\bw_i^{t+1/2}=\bw_i^t-\eta\mathbf g_i^t,
\end{equation}
with $\mathbf g_i^t$ a mini-batch estimate of $\nabla f_i(\bw_i^t)$, and then aggregates with the subset $\Sacc_i\subseteq\N_i$ of its neighbors that survives
filtering,
\begin{equation}\label{eq:aggregate}
\bw_i^{t+1}=\alpha\bw_i^{t+1/2}+(1-\alpha)\sum_{j\in\Sacc_i}W_{ij}\bw_j^{t+1/2},
\end{equation}
with uniform weights $W_{ij}=1/|\Sacc_i|$ or Metropolis weights on the accepted subgraph. The
self-weight $\alpha\in[0,1]$ balances self-reliance against collaboration. We write
$f_i=|\N_i\cap B|$ for the number of Byzantine neighbors of client $i$ and work in the tolerated
regime $f_i<|\N_i|/2$, where honest neighbors form a majority of every neighborhood. In
full-precision similarity-based filtering, $\Sacc_i$ is the set of neighbors whose model lies within the
exact radius
\begin{equation}\label{eq:radius}
\tau_i^t=\gamma e^{-\kappa t/T}\|\bw_i^{t+1/2}\|,
\end{equation}
where $\gamma$ sets the base acceptance radius and $\kappa$ tightens it over the $T$
rounds~\cite{fang2024}. Deciding membership of $\Sacc_i$ requires every neighbor's full model.

\subsection{Count Sketch}
\label{sec:countsketch}

A CS~\cite{charikar2002} compresses a vector by hashing each of its $d$ coordinates into one of
$w$ buckets and adding it there with a random sign. Formally, a hash function $h:[d]\to[w]$ assigns
each coordinate to a bucket and a sign function $\sigma:[d]\to\{\pm1\}$ gives it a random sign, so
bucket $b$ holds
\begin{equation}
\CS(\bw)[b]=\sum_{r:\,h(r)=b}\sigma(r)\,\bw_r,\qquad b=1,\dots,w.
\end{equation}
Coordinates that share a bucket enter with independent random signs, so their cross terms cancel in
expectation. The squared norm of one such table is therefore an unbiased estimate of $\|\bw\|^2$,
with relative variance $O(1/w)$, and a wider table gives a more accurate estimate. The standard
construction repeats this with $R$ independent tables of width $w=O(\varepsilon^{-2})$ and takes the
median of their $R$ estimates, for a total sketch size of $k=Rw$ numbers. The hash and sign functions
of all tables form the seed $\theta=(h_r,\sigma_r)_{r\le R}$. Fixing the seed fixes a linear map
$A_\theta\in\R^{k\times d}$, and client $i$'s sketch is $\bs_i=A_\theta\bw_i$.

This construction makes CS linear, cheap to compute at $O(d)$ with no stored matrix, and
approximately distance-preserving on inputs fixed before the seed. Linearity also means that clients
sharing a seed produce comparable sketches.

\begin{lemma}[CS distance preservation~\cite{ams1999,thorup2012}]\label{lem:cs}
For a pair $\bu,\bv\in\R^d$ fixed independently of $\theta$,
\[
\Pr\big[(1-\varepsilon)\|\bu-\bv\|^2 \le \|A_\theta(\bu-\bv)\|^2 \le (1+\varepsilon)\|\bu-\bv\|^2\big]
\;\ge\; 1-\delta
\]
with a single table of width $k=O(\varepsilon^{-2}\delta^{-1})$ by Chebyshev's inequality, or with
$R=O(\log(1/\delta))$ tables of width $O(\varepsilon^{-2})$ under the median estimator.
\end{lemma}

We write $D_{ij}=\|\bw_i-\bw_j\|$ for the distance between the models of clients $i$ and $j$, and
$\hat d_{ij}=\|\bs_i-\bs_j\|$ for the distance between their sketches. Lemma~\ref{lem:cs} says that
$\hat d_{ij}$ is close to $D_{ij}$ with high probability, which is what makes a sketch usable for
filtering.

The same linearity that makes sketches comparable across clients, however, also makes CS
exploitable as a screen. A linear map from $d$ dimensions down to $k<d$ sends an
entire subspace to zero, its kernel $\ker A$, the set of vectors $\bv$ with $A\bv=\mathbf0$, of
dimension at least $d-k$. Adding a vector from this kernel
changes a model without changing its sketch. Lemma~\ref{lem:cs} does not rule this out, because it
holds only for models chosen independently of the seed, and an adversary that knows the seed can
choose its model to exploit the kernel. Section~\ref{sec:design} builds the protocol around this
observation.

\subsection{Threat model and attacks}
\label{sec:threat}

Byzantine clients may deviate arbitrarily from the protocol, collude, and use Sybil identities, up
to $f_i<|\N_i|/2$ in each neighborhood. Moreover, they know the aggregation rule and the sketch
construction, and because DFL rounds are synchronous, they learn each round's seed at the same moment as
every honest client.

The adversaries we consider are characterized based on timing, namely whether a Byzantine client fixes its model before it knows the seed that defines the round's sketch. A \emph{public-seed} adversary
knows the seed in advance (for example, in scenarios where it never changes), and can therefore tailor its
model to the sketch. A \emph{seed-oblivious} adversary, by contrast, must fix its model before the
seed is drawn. It may still read the full training history and every honest commitment, the binding hash of its
model that each client publishes before the seed is drawn (Section~\ref{sec:protocol}). It can choose
its model from them, but it cannot make that model depend on a seed that does not yet exist. A
client that stops responding once the seed is revealed is simply dropped for the round. Security in
each round therefore rests on two conditions, that every model is fixed before the seed and that
every reported sketch is the true sketch of its sender's model. Section~\ref{sec:security} enforces
both, and because each round draws a fresh seed, the guarantee extends to the whole run.

We evaluate two groups of attacks. The first contains six standard attacks from the literature,
which range from crude disruption to models crafted to slip past distance filters.\medskip

\noindent\textbf{Gaussian noise:} Each Byzantine model is replaced by random noise~\cite{krum2017}.
The result lies far from every honest model and is the easiest attack for a distance filter to
reject.\medskip

\noindent\textbf{Sign-flip:} Each Byzantine client sends a scaled negation of its honest model~\cite{li2019rsa}, reversing descent.\medskip

\noindent\textbf{Directed deviation:} Each Byzantine client moves the honest average against its
direction of progress~\cite{fang2020}, by an amount scaled to the spread of honest models so that
the result still looks plausible.\medskip

\noindent\textbf{ALIE:} ``A little is enough''~\cite{baruch2019} shifts every coordinate of the
honest average by a small multiple of its standard deviation, a change designed to stay inside
distance-based acceptance regions.\medskip

\noindent\textbf{IPM:} Inner-product manipulation~\cite{xie2020} sends a sign-reversed, scaled copy
of the honest average, chosen so that aggregating it reverses the direction of training.\medskip

\noindent\textbf{Null-space:} The model hides a large perturbation in directions the sketch cannot
see. This requires knowing the seed in advance, so only a public-seed
adversary can mount it.\medskip

\noindent The second group contains four attacks that we design against compressed screening itself, to
probe what remains once the null-space attack is closed. A \emph{band-targeted} model sits just
outside the acceptance radius, where sketch noise can occasionally let it through. \emph{Sparse
cancellation} adds two large spikes that disappear from the sketch whenever their coordinates happen
to collide. In \emph{select-after-commit}, a Byzantine client commits a
different candidate to each neighbor and, once the seed is known, sends each neighbor the candidate
that passes its screen. A \emph{forged sketch}, finally, reports a fake sketch, copied from the
receiver's own, for a harmful committed model. Section~\ref{sec:security} analyzes the forged
sketch, and Section~\ref{sec:residual} the other three.

\subsection{Byzantine-robust DFL defenses}
\label{sec:related-defenses}

\begin{table*}[t]
\centering
\small
\caption{Positioning against prior work.}
\label{tab:related-comparison}
\setlength{\tabcolsep}{4pt}
\resizebox{\textwidth}{!}{%
\begin{tabular}{@{}lllllc@{}}
\toprule
Method & Setting & Cost per neighbor before filtering & Check on compressed data & Adaptive failure admitted & Rate shows graph and data effects \\
\midrule
Similarity-based filtering~\cite{guo2022,he2022,fang2024} & Decentralized & $O(d)$, full model & n/a & n/a & Varies by method \\
WFAgg~\cite{cajaraville2025} & Decentralized, dynamic & $O(d)$, full model & n/a & n/a & No guarantee \\
Compression without screening~\cite{qsgd2017,signsgd2018,koloskova2019} & Centralized or decentralized & n/a, no filtering & n/a & n/a & Varies by method \\
FetchSGD~\cite{rothchild2020} & Centralized & n/a, no neighbors & in the sketch domain & Poisoning & n/a, no graph \\
Centralized robust aggregation~\cite{krum2017,gorbunov2022,rammal2024,xia2025} & Centralized & n/a, no neighbors & robust rule at the server & n/a, trusted server & n/a, no graph \\
Robust sketching~\cite{hardt2013,cohen2022,beneliezer2022} & Streaming, one machine & n/a, no neighbors & private randomness & None, trusted machine & n/a, no graph \\
\textbf{SketchGuard} (this paper) & Decentralized, server-free & $\mathbf{O(k)}$, $k\ll d$ & exact, after the fetch & \textbf{None} & \textbf{Yes} \\
\bottomrule
\end{tabular}}
\end{table*}

Similarity-based neighbor filtering is the dominant approach to Byzantine-robust DFL, and the
methods differ mainly in their filtering criteria and aggregation rules. The decentralized setting
is worth defending on its own terms, since Heged{\H{u}}s et al.~\cite{hegedus2021} found it
competitive with server-coordinated FL.\medskip

\noindent\textbf{UBAR}~\cite{guo2022} selects neighbors in two stages, a Euclidean distance test
followed by a loss test, and averages the survivors. It needs neither the Byzantine fraction nor
global connectivity, but it offers no convergence guarantee.\medskip

\noindent\textbf{SCCLIP}~\cite{he2022}, also called ClippedGossip, clips each received model toward
the client's own by a fixed radius. It converges in the non-convex setting without knowledge of the
Byzantine fraction, though the radius must be set conservatively.\medskip

\noindent\textbf{BALANCE}~\cite{fang2024} accepts a neighbor whose model lies within an
exponentially tightening radius. It converges in both the strongly convex and non-convex settings
without knowing the Byzantine fraction or assuming a complete graph, which makes it the only filter
in this family with guarantees in both settings.\medskip

\noindent\textbf{WFAgg}~\cite{cajaraville2025} combines several filters for dynamic and
time-varying topologies but, like UBAR, carries no convergence guarantee.\medskip

\noindent\textbf{Other approaches.} Zeng et al.~\cite{zeng2025poisonproof} weight contributions by
their value to resist poisoning under heterogeneous data. BRFL~\cite{li2025brfl} and
B2DFL~\cite{wang2025b2dfl} anchor trust in a ledger rather than in a distance test, which buys
auditability at the cost of a consensus layer that a bandwidth-limited edge peer must also run.
Krum~\cite{krum2017} and geometric-median aggregation~\cite{pillutla2022} assume a central
aggregator that sees every update. A DFL client, however, sees only its neighborhood, which is why
similarity-based filtering developed as a separate family.\medskip

\noindent Despite their differences, every one of these methods, including BALANCE, receives each neighbor's
full model before it makes any filtering decision.
SketchGuard wraps the family and removes that pre-filtering cost.

\subsection{Compression in federated learning}
\label{sec:related-compression}

Independently of robustness, communication efficiency has been pursued through quantization, as in QSGD~\cite{qsgd2017} and
signSGD~\cite{signsgd2018}, through compressed gossip with convergence under changing
topology~\cite{koloskova2019}, through fewer rounds via control variates~\cite{scaffold2020}, and
through reallocating bandwidth between workers and a server~\cite{han2025owl}. These methods shrink
what is transmitted, and none screens neighbors before aggregating. Consequently, they are orthogonal to screening and compose with it, so we evaluate quantized exchange as an additional baseline rather than as a competing defense.

A second line of work combines compression with Byzantine robustness in server-based
learning~\cite{gorbunov2022,rammal2024,xia2025}. However, these works rely on the server's view of the whole
population, where a robust rule can run before or after decompression. A DFL client has no such
view, so these schemes do not transfer. Sketches have also been used for communication itself.
FetchSGD~\cite{rothchild2020} sketches gradients for server-side aggregation, and
Sketch-fusion~\cite{dai2024sketchfusion} sketches gradients for communication-efficient distributed
training. Both sketch what is transmitted. SketchGuard, by contrast, sketches what is compared, and the difference
matters, because a decision taken in the sketch domain is what an adaptive adversary can steer.
Existing literature surveys identify communication efficiency and adversarial robustness as separate open
problems~\cite{kairouz2021,lim2020}. This paper proposes a solution at their intersection.

\subsection{Adaptive robustness of sketches}
\label{sec:related-sketch}

Screening on sketches also inherits a known limitation of sketching itself. Hardt and
Woodruff~\cite{hardt2013} proved that no linear sketch preserves Euclidean norms on
adaptively chosen inputs, and Cohen et al.~\cite{cohen2022} specialized the result to CS.
Ben-Eliezer et al.~\cite{beneliezer2022} restored robustness by making the sketch randomness
unpredictable to the adversary. SketchGuard builds on these prior results, carrying them into decentralized Byzantine screening. This
is non-trivial, because in the DFL setting the weakness of linear sketches surfaces as two concrete
attacks, poisoning and bandwidth amplification. The repair must also change, because robust streaming
relies on randomness held privately by one trusted machine, and a Byzantine peer-to-peer graph has no
such machine.
Table~\ref{tab:related-comparison} summarizes how SketchGuard relates to each of these lines of work.

%% file: sections_jpdc/design.tex
\section{SketchGuard}
\label{sec:design}

In every similarity-based filter, the communication cost and the filtering decision are coupled
by design. A client must receive a neighbor's full model before it can compute the distance that
decides whether to accept it. The coupling, however, is not logically necessary. The decision needs only a
distance that is approximately preserved, which a sketch supplies, while full fidelity is needed
only by the aggregation that follows. SketchGuard accordingly breaks the coupling by splitting each round into
a cheap sketch exchange used for screening and a selective fetch restricted to accepted neighbors.
A rejected neighbor then costs $O(k)$ in communication rather than $O(d)$.

However, the side effect of breaking the coupling is that it moves the decision onto an object the adversary can steer, since the
sketch is linear and its kernel is large. Three rules keep the decoupling safe and cheap. First, the
sketch decides only which models are fetched. Every fetched model is checked against the wrapped
filter's exact radius $\tau_i^t$ before it enters the aggregate, so safety never depends on the
sketch. Second, each client commits to its model before the round's seed exists, so no model can be
built against the map. Third, a sender whose opened model does not reproduce the sketch it sent is
banned, which binds the sketch to the committed model. Section~\ref{sec:security} shows that each
rule closes an independent failure.

\subsection{Protocol description}
\label{sec:protocol}

\input{sections_jpdc/fig_protocol}

Each client runs the same round with public parameters, the filter's radius $\gamma$ and decay
$\kappa$, the self-weight $\alpha$, the sketch size $k$, and the learning rate $\eta$. At round $t$,
client $i$ executes the six phases below. Figure~\ref{fig:protocol} shows their high-level flow, and
the list states each step precisely.

\begin{enumerate}
\item\label{ph:train} \textit{Local training.} Client $i$ takes the gradient step of
\eqref{eq:local-update} on its own data and obtains $\bw_i^{t+1/2}$.
\item\label{ph:commit} \textit{Commitment.} Client $i$ draws a private nonce $r_i$ and sends the
commitment $c_i=H(\bw_i^{t+1/2}\|r_i)$, a SHA-256 digest of its serialized model and the nonce, to
every neighbor $j\in\N_i$, and receives $c_j$ in return. The commitment is binding, so the model is
fixed from this point on, and hiding, so it reveals nothing about the model. Anyone who later
receives the pair $(\bw_j,r_j)$ can check it against $c_j$, an operation we call \emph{opening} the
commitment. Since the protocol assumes no authenticated broadcast, each commitment travels on its own
edge, and Section~\ref{sec:residual} bounds what a sender gains from that freedom.
\item\label{ph:sketch} \textit{Seed and sketch exchange.} Only after every commitment has been exchanged is
the round seed $\theta_t$ fixed, read from a public randomness beacon~\cite{drand2020} or derived from
a commit-reveal of client nonces. Client $i$ computes its sketch $\bs_i=\CS_{\theta_t}(\bw_i^{t+1/2})
\in\R^k$, sends it to its neighbors, and receives theirs, $O(k|\N_i|)$ values in total.
\item\label{ph:screen} \textit{Sketch-domain screening.} Client $i$ applies the wrapped filter's rule
to sketches instead of models. It sets the sketch-domain radius $\hat\tau_i^t=\gamma
e^{-\kappa t/T}\|\bs_i\|$ and keeps the neighbors whose sketch lies within it,
\begin{equation}\label{eq:filter}
\Sacc_i^t=\big\{j\in\N_i:\ \|\bs_i-\bs_j\|\le\hat\tau_i^t\big\}.
\end{equation}
The exponential decay reflects convergence. As honest models align, the acceptable spread among
them narrows.
\item\label{ph:verify} \textit{Fetch and verification.} Client $i$ requests the full model and nonce
of every $j\in\Sacc_i^t$, either from $j$ itself or from any peer that holds a copy matching $c_j$. It
then drops $j$ for the round if any of four checks fails, namely if the fetch times out, if
$(\bw_j,r_j)$ does not open $c_j$, if the re-sketch $\CS_{\theta_t}(\bw_j)$ differs from the reported
$\bs_j$, or if $\bw_j$ lies outside the wrapped filter's exact radius,
$\|\bw_i^{t+1/2}-\bw_j\|>\tau_i^t$ with $\tau_i^t$ from \eqref{eq:radius}. A delivery that fails to open its
commitment, or that opens it but does not reproduce the sketch its owner reported, proves
misbehavior, because an honest client never produces either. The receiver therefore bans the peer
that delivered it for the rest of training, namely the relay if the model came through one, after
which it fetches directly, and otherwise the neighbor itself.
\item\label{ph:agg} \textit{Aggregation.} Client $i$ combines its own model with the neighbors that
passed every check, using \eqref{eq:aggregate}, by default with uniform weights,
$\bw_i^{t+1}=\alpha\bw_i^{t+1/2}+\frac{1-\alpha}{|\Sacc_i^t|}\sum_{j\in\Sacc_i^t}\bw_j$, where
$\Sacc_i^t$ now denotes the verified set. If no neighbor survives, the client keeps its own model.
\end{enumerate}

\noindent Across all six phases, dropping or banning a neighbor needs no agreement from any other node.
Instead, each client decides its own accepted set and proceeds, with per-neighbor timeouts bounding
how long it waits on a fetch. Relative to a full-precision filter, SketchGuard adds the commitment
and the seed draw (Phases~\ref{ph:commit} and~\ref{ph:sketch}), the sketch exchange and screen
(Phases~\ref{ph:sketch} and~\ref{ph:screen}), and the checks and ban of Phase~\ref{ph:verify}, which
cost $O(d)$ local compute per fetched model and no extra communication.

\subsection{Security of compressed screening}
\label{sec:security}

A round of SketchGuard gives the adversary three chances to act on something the defender relies
on. It can send a model whose sketch looks acceptable but whose full model is harmful, it can choose
its model after learning how sketches are computed, and it can choose the sketch it reports after
the seed is revealed. The first
concerns safety, meaning what can reach the aggregate, and the other two concern cost, meaning how
many useless downloads the adversary can force. Re-sketching a fetched model and opening its
commitment already stop the simplest attack, a benign sketch followed by a different model, because
both checks are deterministic functions of the model. Except where noted, every result holds for
an arbitrary linear sketch, not only for CS. Proofs
of the results in this subsection and in Section~\ref{sec:residual} are in \ref{app:security}.

\paragraph{Rule 1, trust only the exact test.}
After fetching a model, the client applies the wrapped filter's own test to the full model
(Phase~\ref{ph:verify}). A model therefore enters the aggregate only if the uncompressed filter
would have accepted it, regardless of what its sketch said. Lemma~\ref{lem:exact} turns this into a bound on how far
the accepted neighbors can move a client's model.

\begin{lemma}[Exact verification bounds the displacement]\label{lem:exact}
Let $S_i$ be the set client $i$ aggregates, with weights $W_{ij}\ge0$, and suppose every $j\in S_i$
satisfies $D_{ij}\le\tau$. Then
\[
\Big\|\sum_{j\in S_i} W_{ij}(\bw_j-\bw_i)\Big\| \le \Big(\sum_{j\in S_i} W_{ij}\Big)\tau ,
\]
and the part contributed by the accepted Byzantine neighbors $B_i\subseteq S_i$ is at most
$b_i\tau$, where $b_i=\sum_{j\in B_i}W_{ij}$.
\end{lemma}

\noindent It is the triangle inequality applied to a non-negative combination, and the same chain restricted to $B_i$ gives the Byzantine part. The bound never mentions the sketch. Compression can therefore cost bandwidth, but it cannot weaken
the wrapped filter's safety. The same holds for the average of the honest clients, because
symmetric mixing moves that average only through the Byzantine contributions.

\begin{lemma}[Symmetric mixing preserves the honest mean]\label{lem:mean}
If $W$ is symmetric on a finite index set, then $\sum_i\sum_j W_{ij}(\bw_j-\bw_i)=0$.
\end{lemma}

\noindent Exchanging the two index names and using $W_{ji}=W_{ij}$ makes the two double sums equal. Under Metropolis weights on mutually accepted honest edges, $W$ is
symmetric among the honest clients, so their exchanges with each other cancel in the honest average.
Only the accepted-Byzantine term of Lemma~\ref{lem:exact} remains, and it is the quantity $\beta$ on
which the convergence analysis of Section~\ref{sec:convergence} rests.\medskip

With safety settled, the sketch decides only which models are downloaded, and it can err in two
ways. It can skip an honest neighbor that the exact test would accept, which costs completeness, or
admit a Byzantine model that the exact test then rejects, which costs a wasted fetch. Both errors
require the sketch to misjudge a distance. Recall that the screen accepts $j$ when
$\hat d_{ij}\le\hat\tau_i^t=\gamma e^{-\kappa t/T}\|\bs_i\|$, and the exact test when
$D_{ij}\le\tau_i^t$. We say the sketch is \emph{accurate} for the pair $(i,j)$, an event we write
$E_\varepsilon(i,j)$, when it estimates both the squared distance $D_{ij}^2$ and the client's own
squared norm $\|\bw_i\|^2$ to within a factor $1\pm\varepsilon$. Lemma~\ref{lem:cs}, applied once
to each estimate, makes the sketch accurate with probability at least $1-2\delta$.

An accurate sketch can err only near the acceptance radius. Two radii mark the region where it can,
\begin{equation}\label{eq:band}
\gin=\gamma\sqrt{\frac{1-\varepsilon}{1+\varepsilon}},\qquad
\geff=\gamma\sqrt{\frac{1+\varepsilon}{1-\varepsilon}},
\end{equation}
and we call the interval between $\gin e^{-\kappa t/T}\|\bw_i\|$ and $\geff e^{-\kappa t/T}\|\bw_i\|$
the \emph{ambiguous band}.

\begin{lemma}[Screening reduction]\label{lem:screen}
Suppose the sketch is accurate for $(i,j)$. (a) If the screen accepts $j$, then
$D_{ij}\le\geff e^{-\kappa t/T}\|\bw_i\|$. (b) If $D_{ij}\le\gin e^{-\kappa t/T}\|\bw_i\|$, the
screen accepts $j$.
\end{lemma}

The ambiguous band is therefore the only place where sketch noise, rather than an adversary, can flip
a decision. This guarantee, however, rests on Lemma~\ref{lem:cs}, which holds only for a
model chosen without knowledge of the seed. We call such a model seed-oblivious.

\begin{definition}[Seed-oblivious model]\label{def:oblivious}
A delivered model $\bw_j$ is \emph{seed-oblivious} if it is fixed before, and independently of, the
round seed $\theta_t$, that is, if it is measurable with respect to a $\sigma$-algebra
$\mathcal F$ independent of $\theta_t$.
\end{definition}

\noindent The next theorem states what sketch noise costs for such models.

\begin{theorem}[What a sketch failure costs]\label{thm:cost}
Let $j$ be seed-oblivious (Definition~\ref{def:oblivious}). (i) A wasted fetch requires
$\tau_i^t<D_{ij}\le\geff e^{-\kappa t/T}\|\bw_i\|$, that is, a model inside the ambiguous band.
(ii) A model beyond the band is fetched with probability at most $2\delta$. (iii) An honest model
within $\gin e^{-\kappa t/T}\|\bw_i\|$ is skipped with probability at most $2\delta$.
\end{theorem}

The failure probability $\delta$ thus governs bandwidth and connectivity, never safety. A single
table of width $k$ gives $\delta=O(1/(k\varepsilon^2))$, and the median of several independent
tables drives it down exponentially.

\begin{lemma}[Median of $R$ tables]\label{lem:median}
If more than half of $R$ per-table estimates lie within a factor $1\pm\varepsilon$ of the true
squared distance, so does their median. If each table fails independently with probability $p<1/2$,
the median fails with probability at most $\exp(-2R(1/2-p)^2)$.
\end{lemma}

\paragraph{Rule 2, commit before the seed.}
Theorem~\ref{thm:cost} assumes that the Byzantine model was fixed before the seed. An adversary that
knows the seed in advance is not so constrained, and it can exploit the kernel of the sketch map
described in Section~\ref{sec:countsketch}.

\begin{theorem}[Public seed, poisoning or amplification]\label{thm:public}
Suppose the adversary knows the sketch map $A\in\R^{k\times d}$ with $k<d$. For every model $\bw$ and
every radius $\tau$, it can build a model $\tilde\bw$ with the same sketch, $A\tilde\bw=A\bw$, but
with $\|\tilde\bw-\bw\|>\tau$.
\end{theorem}

The model $\tilde\bw$ passes every sketch screen and the re-sketch check, because its sketch is
exactly that of a benign model, and it fails the exact test. The outcome depends on Rule~1. Without
the exact test, the aggregate is poisoned by an arbitrary amount. With it, the aggregate is safe, but
every such neighbor forces one full download in every round, which gives back the saving that
motivated compression. For CS the construction is also cheap. Within one table the rows
have disjoint supports, so the projection onto $\ker A$ is available in closed form at $O(d)$ cost,
and across $R$ tables a few sweeps of alternating projections converge to the joint kernel at
$O(Rd)$ per sweep.

The attack requires the adversary to choose $\tilde\bw$ after learning the sketch map, so fixing the
order of revelation removes it, by forcing every model to be seed-oblivious in the sense of
Definition~\ref{def:oblivious}. The following assumption states what that requires.

\begin{assumption}[Binding commitment, unpredictable seed]\label{as:commit}
The commitment is computationally binding, and the seed $\theta_t$ is drawn after the commitment
barrier, independently of everything committed before it.
\end{assumption}

\begin{theorem}[Commit-then-sketch restores efficiency]\label{thm:cts}
Under Assumption~\ref{as:commit}, every delivered model is seed-oblivious. If moreover every reported
sketch is the true sketch of its sender's model, a Byzantine neighbor outside the ambiguous band
wastes a fetch with probability at most $2\delta$ per round, and one inside the band wastes at most
one fetch per round.
\end{theorem}

Safety is unchanged, since it rests on Lemma~\ref{lem:exact} alone. Rules~1 and~2 together
therefore make screening both safe and cheap, whereas Rule~1 alone is safe but defeatable in
bandwidth. A client that withholds its commit-reveal opening can bias a redraw of the seed, but it
cannot reopen Theorem~\ref{thm:public}, because its model was already fixed.

\paragraph{Rule 3, ban on mismatch.}
Theorem~\ref{thm:cts} assumes that each reported sketch is honest, and the commitment alone does not
guarantee this, because the sketch is sent after the seed. A Byzantine sender can commit a harmful
model and report a sketch copied from the receiver's own, which passes every screen. Left unchecked,
this forces a wasted fetch on every Byzantine edge in every round, the amplification of
Theorem~\ref{thm:public} regained without knowing the seed in advance. The fetch itself exposes the
forgery. The delivered model must open its commitment, and re-sketching it then reveals the
mismatch, while a payload that fails to open is equally revealing. An honest client never produces
either, so the receiver bans the peer that delivered it (Phase~\ref{ph:verify}).

\begin{theorem}[Ban on mismatch]\label{thm:ban}
An honest sender is never banned. A forged sketch, or a delivery that fails to open its commitment,
wastes at most one fetch per edge over the whole run, direct or through a relay, and therefore at
most $|E_B|$ fetches over any set $E_B$ of Byzantine edges.
\end{theorem}

After a sender's first forgery, every sketch a client screens on is therefore the true sketch of its
sender's model, which is the hypothesis of Theorem~\ref{thm:cts}. Sketches must be computed over the
transmitted payload, so that quantized exchange cannot make an honest sketch mismatch.

\subsection{Residual seed-oblivious attacks}
\label{sec:residual}

The three rules close the null-space attack and the forged sketch, but sketch noise remains, and an
adversary that cannot see the seed can still try to exploit it. We identify three such attacks.
None of them affects safety, which rests on Lemma~\ref{lem:exact}, and each costs a bounded number
of wasted fetches.

\paragraph{Sparse cancellation.}
Adding two large spikes of opposite sign on two coordinates leaves a model's sketch unchanged
whenever the two coordinates happen to fall into the same bucket with the same sign. This needs no
knowledge of the seed.

\begin{theorem}[Sparse cancellation]\label{thm:sparse}
For a single CS table of width $k$, adding $c(\mathbf e_a-\mathbf e_b)$ with $a\neq b$ leaves a
model's sketch unchanged exactly when $h(a)=h(b)$ and $\sigma(a)=\sigma(b)$, and otherwise moves some
bucket by at least $|c|$. Under uniform independent hashing and signs, cancellation occurs with
probability $1/(2k)$, whatever the seed.
\end{theorem}

A cancelled spike of any size passes a single-table screen, so commit-then-sketch cannot stop it.
The exact test does, rejecting the spike after one wasted fetch per success, a rate of $1/800$ at
$k=400$. A median over $R$ tables lowers that rate further, because the spike hides from the median
only if at least $\lceil R/2\rceil$ tables cancel at once, which happens with probability at most
$\binom{R}{\lceil R/2\rceil}(2k)^{-\lceil R/2\rceil}$ (\ref{app:security}). The median therefore
changes bandwidth, and the exact test alone provides safety.

\paragraph{Band-targeted placement.}
An adversary that knows $\gamma$, $\kappa$, and $\varepsilon$ can place its committed model just
outside the radius it estimates from its honest neighbors, so that some receivers see it inside the
ambiguous band. By Theorem~\ref{thm:cost}, such a model wastes a fetch with probability at most
$2\delta$ outside the band and at most once per round inside it, and the exact test rejects it once
$D_{ij}>\tau_i^t$. A model placed inside the radius, by contrast, is accepted by the wrapped filter
itself, and the sketch cannot change that.

\paragraph{Equivocation.}
Because each commitment travels on its own edge (Phase~\ref{ph:commit}), nothing exposes an identity
that commits different models to different neighbors. Such an identity can therefore
commit one candidate per edge and, once the seed is revealed, forward to each neighbor the candidate
that passes that neighbor's screen. This select-after-commit strategy forces at most one wasted fetch
per Byzantine edge per round and has no effect on safety.

\subsection{Complexity analysis}
\label{sec:complexity}

Consider one honest node per round with $N=|\N_i|$ neighbors, a model payload of $m$ bytes
($m=4d$ in float32), a sketch payload of $s=4kR+32$ bytes ($R$ tables of $k$ counters plus a
commitment digest), and $A=|\Sacc_i|$ accepted and fetched neighbors. Table~\ref{tab:bytes} gives
the per-design cost. Each row follows from counting the bytes a design sends and receives. Comparing the first two rows, the fraction of the bytes SketchGuard moves is
\begin{equation}\label{eq:ratio}
\frac{Ns+Am}{Nm}=\frac{s}{m}+\frac{A}{N}.
\end{equation}
At $d=833{,}967$ and $k=400$, $s/m\approx0.0005$, so the ratio is essentially the
acceptance rate. The saving scales with the number of rejected neighbors and with the model size,
and it is capped at one half under honest majority, where at least half the neighbors are accepted
and fetched regardless. The amplification row is Theorem~\ref{thm:public} in bytes ($A=N$). It costs
exactly $Ns$ more than full exchange. Computation, however, does not follow the same pattern. Screening costs
$O(k|\N_i|)$ rather than $O(d|\N_i|)$, but every accepted neighbor then costs a commitment opening,
a re-sketch, and the exact test, each $O(d)$, where BALANCE pays one distance test. SketchGuard
therefore spends more local compute than BALANCE whenever a sizable fraction of neighbors is
accepted, and Section~\ref{sec:cost} measures by how much.

\begin{table}[t]
\centering
\caption{Per-node, per-round communication bytes.}
\label{tab:bytes}
\resizebox{\columnwidth}{!}{%
\begin{tabular}{ll}
\toprule
Design & Bytes \\
\midrule
BALANCE (full exchange) & $N m$ \\
SketchGuard, any seed & $N s + A m$ \\
Public seed, exact verif.\ (amplified) & $N s + N m \ge N m$ \\
Saving of screening & $(N-A)m - N s$ \\
Honest majority ($A\ge N/2$) & saving $\le N m / 2$ \\
Quantized fetch, $m'\le m$ & saves further $A(m-m')$ \\
Relay fetch, $\rho$ requesters per model & uploader sends $m$, not $\rho m$ \\
\bottomrule
\end{tabular}}
\end{table}

\paragraph{Composition with quantization.}
Screening also composes with quantization, because the two act on different phases. Reducing the fetched payload from
$m$ to a quantized $m'\le m$, for instance $m'=d+4\lceil d/1024\rceil$ at 8 bits, reduces the cost
by exactly $A(m-m')$ however $A$ was determined, and leaves the sketch overhead $Ns$ unchanged.

\paragraph{Content-addressed relay fetch.}
The commitment also has a second use on the continuum. Because $c_j$ binds $\bw_j$ before any
fetch, the model is content-addressed, so any peer that holds a copy can serve it, not just its
owner. The receiver then runs the same checks as for a direct
fetch. A malicious relay can only withhold the model, which the timeout in Phase~\ref{ph:verify} handles, and cannot substitute one, so safety is unchanged. On the continuum, this lets cloud peers
act as verified caches for edge peers. Each edge node $j$ has a designated cloud neighbor, the one
with the lowest identifier, that fetches $j$ directly, and every other requester adjacent to it
fetches $j$'s model from that relay, so an edge uplink carries each model once per round in the best
case, the last row of Table~\ref{tab:bytes}. If a designated relay withholds the model, its requesters wait one timeout and then fetch from $j$
directly.

%% file: sections_jpdc/fig_protocol.tex
\begin{figure*}[t]\centering
\resizebox{0.95\textwidth}{!}{%
\begin{tikzpicture}[>=Stealth, font=\small,
  lab/.style={font=\footnotesize\bfseries, align=center},
  capt/.style={font=\scriptsize, text=black!75, align=center},
  flow/.style={->, line width=1.3pt, gray!65!black}, ta/.style={->, semithick}]

\def\dx{3.15}
\foreach \i in {0,...,5} \coordinate (c\i) at (\i*\dx,0);

\node[lab] at ($(c0)+(0,1.25)$) {Phase 1\\ Local training};
\node[lab] at ($(c1)+(0,1.25)$) {Phase 2\\ Commit};
\node[lab] at ($(c2)+(0,1.25)$) {Phase 3\\ Sketch \& exchange};
\node[lab] at ($(c3)+(0,1.25)$) {Phase 4\\ Screen};
\node[lab] at ($(c4)+(0,1.25)$) {Phase 5\\ Fetch \& verify};
\node[lab] at ($(c5)+(0,1.25)$) {Phase 6\\ Aggregate};

\foreach \i in {0,...,5} {\clientnode{p\i}{(c\i)}}
\foreach \a/\b in {0/1,1/2,2/3,3/4,4/5} {\draw[flow] (p\a)--(p\b);}

% 1: gears
\node[star,star points=9,star point ratio=0.82,draw=gray!55,fill=gray!30,minimum size=0.6cm]
      at ($(c0)+(-0.17,-1.5)$) {};
\node[star,star points=9,star point ratio=0.82,draw=gray!55,fill=gray!30,minimum size=0.44cm]
      at ($(c0)+(0.22,-1.74)$) {};
\node[capt] at ($(c0)+(0,-2.30)$) {gradient step};

% 2: padlock (commit)
\begin{scope}[shift={($(c1)+(0,-1.5)$)}]
  \draw[thick,fill=blue!15,rounded corners=1pt] (-0.23,-0.23) rectangle (0.23,0.15);
  \draw[thick] (-0.13,0.15) arc(180:0:0.13);
  \fill (0,-0.04) circle (0.03);
\end{scope}
\node[capt] at ($(c1)+(0,-2.30)$) {$c_i{=}H(w_i\|r_i)$};

% 3: exchange sketches with neighbours (one Byzantine)
\honnode{n3a}{($(c2)+(-0.85,-1.7)$)}
\byznode{n3b}{($(c2)+(0.85,-1.7)$)}
\draw[<->,green!45!black,semithick] (p2.south west) to[out=-115,in=70] (n3a);
\draw[<->,green!45!black,semithick] (p2.south east) to[out=-65,in=110] (n3b);
\sketchsq{($(c2)+(-0.5,-1.02)$)}{green!60!black}
\sketchsq{($(c2)+(0.5,-1.02)$)}{green!60!black}
\node[capt] at ($(c2)+(0,-2.30)$) {exchange sketches\\ ($O(k)$)};

% 4: screen -- accept close (check), reject far Byzantine (cross)
\sketchsq{($(c3)+(-0.6,-1.55)$)}{green!60!black} \okmark{($(c3)+(-0.6,-1.08)$)}
\sketchsq{($(c3)+(0.0,-1.55)$)}{green!60!black}  \okmark{($(c3)+(0.0,-1.08)$)}
\sketchsq{($(c3)+(0.6,-1.55)$)}{red!70}          \xmark{($(c3)+(0.6,-1.08)$)}
\node[capt] at ($(c3)+(0,-2.30)$) {keep if\\ $\|s_i{-}s_j\|\le\hat\tau$};

% 5: fetch FULL models from accepted only + verify
\honnode{n5a}{($(c4)+(-0.85,-1.72)$)}
\honnode{n5b}{($(c4)+(0.85,-1.72)$)}
\draw[->,blue!55!black,semithick] (n5a) to[out=68,in=-112] (p4.south west);
\draw[->,blue!55!black,semithick] (n5b) to[out=112,in=-68] (p4.south east);
\modelsq{($(c4)+(-0.52,-1.08)$)}
\modelsq{($(c4)+(0.52,-1.08)$)}
\okmark{($(c4)+(0,-1.08)$)}
\node[capt] at ($(c4)+(0,-2.30)$) {fetch, exact-verify ($O(d)$),\\ ban on mismatch};

% 6: aggregate
\modelsq{($(c5)+(-0.45,-1.5)$)} \node at ($(c5)+(0,-1.5)$) {\small$+$};
\modelsq{($(c5)+(0.45,-1.5)$)}  \node[capt] at ($(c5)+(0.92,-1.5)$) {$\cdots$};
\node[capt] at ($(c5)+(0,-2.30)$) {$\alpha w_i{+}(1{-}\alpha)\bar{w}$};

% beacon: commit -> beacon -> sketch (order shown by the two arrows)
\node[star,star points=6,star point ratio=0.5,draw=orange!80!black,fill=orange!35,thick,
      minimum size=0.72cm] (bc) at ($(c1)!0.5!(c2)+(0,2.25)$) {\scriptsize$\theta_t$};
\node[capt] at ($(bc)+(0,0.80)$) {public beacon};
\draw[ta,dashed,black!50] ($(c1)+(0,1.8)$) to[out=90,in=200] (bc.west);
\draw[ta,orange!75!black,thick] (bc.east) to[out=-20,in=90] ($(c2)+(0,1.8)$);

% round loop-back, routed outside the columns
\draw[flow,rounded corners=7pt]
  (p5.east) -- ($(c5)+(1.25,0)$) -- ($(c5)+(1.25,-2.95)$)
  -- ($(c0)+(-1.25,-2.95)$)
  -- ($(c0)+(-1.25,0)$) -- (p0.west);
\end{tikzpicture}}
\caption{One round of SketchGuard.}
\label{fig:protocol}
\end{figure*}

%% file: sections_jpdc/convergence.tex
\section{Convergence analysis}
\label{sec:convergence}

We show that SketchGuard keeps the convergence guarantees of the filter it wraps, even though
screening happens in the compressed domain. The proof strategy has three steps. We first use
Lemma~\ref{lem:exact} to bound how far the accepted neighbors can move a client's model in one
round. We then treat that bounded move as a perturbation of gradient descent and derive a one-step
descent inequality. Finally we apply the Polyak--{\L}ojasiewicz or non-convex argument, and extend
it to the honest average with explicit topology and heterogeneity terms. Notably, the sketch enters none of
these steps, because exact verification removes it from the path to the aggregate. Full proofs are in
\ref{app:convergence}.

\subsection{Technical assumptions}
\label{sec:assumptions}

\begin{assumption}[Smoothness and strong convexity]\label{as:sc}
$F$ and every honest $f_i$ are $L$-smooth, and $F$ is $\mu$-strongly convex where
Theorem~\ref{thm:pl} requires it.
\end{assumption}
\begin{assumption}[Bounded variance]\label{as:var}
Each honest stochastic gradient satisfies $\mathbb{E}[\mathbf g_i]=\nabla f_i(\bw_i)$ and
$\mathbb{E}\|\mathbf g_i-\nabla f_i(\bw_i)\|^2\le\sigma^2$.
\end{assumption}
\begin{assumption}[Bounded iterates]\label{as:bd}
For honest $i$, $\|\bw_i\|\le\psi$ and $\|\nabla F(\bw_i)\|\le\rho$.
\end{assumption}
\begin{assumption}[Bounded heterogeneity]\label{as:het}
For every honest $i$ and every $\bw$, $\|\nabla f_i(\bw)-\nabla F(\bw)\|\le\zeta$, with $\zeta=0$
recovering the IID case.
\end{assumption}
\begin{assumption}[Honest connectivity and mixing]\label{as:mix}
The subgraph of mutually accepted honest edges is connected in every round. Its Metropolis weights,
$[W_t]_{ij}=1/(1+\max(\deg_i,\deg_j))$ for $j\in\Sacc_i^t$ and $i\in\Sacc_j^t$ with $[W_t]_{ii}$
completing the row, give a symmetric doubly stochastic matrix with uniform spectral bound
$\lambda=\sup_t\|W_t-\tfrac1{|H|}\mathbf1\mathbf1^\top\|_2<1$.
\end{assumption}

Assumptions~\ref{as:sc} to~\ref{as:bd} are standard in the Byzantine DFL
literature~\cite{fang2024,he2022}. Assumption~\ref{as:var} bounds the noise of the local step. It is
not needed by Theorems~\ref{thm:onestep} to~\ref{thm:nc}, which are stated for a deterministic error
bound, and enters only through the consensus recursion of Theorem~\ref{thm:topo}.
Assumption~\ref{as:het} is the pointwise form of the averaged non-IID measure
of~\cite{fang2024}, which it implies. The per-client theorems need the pointwise form, whereas the
honest-average theorem uses it only through the consensus gap. Because the acceptance radius is node-local,
acceptance can be one-directional. Assumption~\ref{as:mix} therefore conservatively keeps only mutually
accepted honest edges, and accepted Byzantine contributions enter instead through the displacement
bound below.

\subsection{Bounding the displacement}
\label{sec:round-model}

Screening enters the analysis at exactly one place, the displacement that aggregation adds to a
client's own descent step. We write one round as
\begin{equation}\label{eq:round}
x_{t+1}=x_t-\eta g_t+\Delta_t ,
\end{equation}
where $x_t$ is an honest client's model $\bw_i^t$ (Theorems~\ref{thm:onestep} to~\ref{thm:nc}) or the
honest average $\bar\bw^t$ (Theorem~\ref{thm:topo}), $g_t$ is the direction followed, and $\Delta_t$
is what aggregation adds. Two scalars bound the error in each. The gradient error
$\|g_t-\nabla F(x_t)\|\le\zeta$ comes from Assumption~\ref{as:het} for a single client and from the
consensus gap for the honest average. The displacement $\|\Delta_t\|\le\beta$ comes from
Lemma~\ref{lem:exact}, as $\beta=(1-\alpha)\tau_{\max}$ for a single client under uniform weights,
which conservatively charges the pull toward accepted honest neighbors as error, and as
$\beta=b\,\tau_{\max}$ for the Metropolis honest average, where the honest part cancels by
Lemma~\ref{lem:mean} and $b$ is the accepted-Byzantine weight. Both rest on
$\tau_{\max}\le\gamma\psi$, the exact radius, so the sketch radius $\geff$ never appears in
$\beta$.

\subsection{Main convergence results}
\label{sec:rates}

\begin{theorem}[One-step descent]\label{thm:onestep}
Suppose $F$ is $L$-smooth, $\|\nabla F(x)\|\le\rho$, and $\eta L\le 1/8$. For any $g_t$ and $\Delta$
with $\|g_t-\nabla F(x)\|\le\zeta$ and $\|\Delta\|\le\beta$,
\[
F(x-\eta g_t+\Delta) \le F(x) - \frac\eta4\|\nabla F(x)\|^2 + \frac{3\eta}4\zeta^2 + \rho\beta +
L\beta^2 .
\]
\end{theorem}
\begin{theorem}[Strongly convex rate]\label{thm:pl}
Under the hypotheses of Theorem~\ref{thm:onestep} in every round, suppose $F$ also satisfies the
Polyak--{\L}ojasiewicz inequality $\|\nabla F(x)\|^2\ge 2\mu(F(x)-F^\star)$, which
$\mu$-strong convexity implies, with $\mu\le L$. Then along \eqref{eq:round}, with
$C=\frac{3\eta}4\zeta^2+\rho\beta+L\beta^2$,
\[
F(x_T)-F^\star \le \Big(1-\frac{\mu\eta}2\Big)^T\big(F(x_0)-F^\star\big) + \frac{2C}{\mu\eta},
\]
a floor of $\frac{3\zeta^2}{2\mu}+\frac{2(\rho\beta+L\beta^2)}{\mu\eta}$.
\end{theorem}
\begin{theorem}[Non-convex rate]\label{thm:nc}
Under the hypotheses of Theorem~\ref{thm:onestep} in every round, along \eqref{eq:round},
\[
\frac1T\sum_{t<T}\|\nabla F(x_t)\|^2 \le \frac{4\big(F(x_0)-F^\star\big)}{\eta T} + 3\zeta^2 +
\frac{4(\rho\beta+L\beta^2)}{\eta} .
\]
\end{theorem}
\noindent Theorems~\ref{thm:pl} and~\ref{thm:nc}, however, bound only one honest client's iterate. The next
result bounds
the network, through the honest average $\bar\bw^t=\frac1{|H|}\sum_{i\in H}\bw_i^t$ and its
consensus error $\Omega_t=\frac1{|H|}\sum_{i\in H}\|\bw_i^t-\bar\bw^t\|^2$.

\begin{theorem}[Topology and heterogeneity]\label{thm:topo}
Apply Theorem~\ref{thm:nc} to the Metropolis honest average $x_t=\bar\bw^t$ with
$g_t=\frac1{|H|}\sum_{i\in H}\nabla f_i(\bw_i^t)$, for which the gradient error is at most
$L\sqrt{\Omega_t}$, and suppose the consensus recursion $\Omega_{t+1}\le\lambda^2\Omega_t+B_\xi$
holds with $B_\xi=O\big(\eta^2(\sigma^2+\zeta^2)+\beta^2\big)$, with coefficients as
in~\cite{koloskova2020}. Then
\begin{multline*}
\frac1T\sum_{t<T}\mathbb E\|\nabla F(\bar\bw^t)\|^2 \le
\frac{4\big(F(\bar\bw^0)-F^\star\big)}{\eta T} \\
+ \frac{3L^2\Omega_0}{T(1-\lambda^2)} + \frac{3L^2 B_\xi}{1-\lambda^2} +
\frac{4(\rho\beta+L\beta^2)}\eta .
\end{multline*}
\end{theorem}
Each bound has the same shape. The first term is the initialization gap, the $\zeta^2$ terms carry
data heterogeneity, and $\rho\beta+L\beta^2$ carries the aggregation displacement.
Theorem~\ref{thm:topo} adds a consensus term growing as $1/(1-\lambda^2)$, so sparse or poorly
mixing honest graphs pay more, and it reduces to Theorem~\ref{thm:nc} as $\lambda,\zeta\to0$. The
displacement term, however, carries no factor of $\eta$, because the acceptance radius scales
with the model norm rather than the step size, so it survives as a floor. That floor scales with the
accepted-Byzantine weight $b$ and vanishes when no Byzantine model is accepted.

Compression never loosens the displacement bound. A bound proved in the sketch domain would carry $\geff$
in place of the exact radius and would be void against a public-seed adversary
(Theorem~\ref{thm:public}), whereas exact verification leaves the bounds independent of $k$,
$\varepsilon$, and $\delta$. These enter only through completeness, since a sketch that skips an
honest neighbor can thin the honest subgraph of Assumption~\ref{as:mix}.

\subsection{Scope of the guarantee}
\label{sec:scope}

Because they rely only on the exact radius, the theorems hold for any filter with an exact acceptance
radius. SketchGuard adds nothing to and
removes nothing from the wrapped filter's convergence, since its aggregated set satisfies the same
radius condition by Lemma~\ref{lem:exact}. The guarantee, however, is
convergence-theoretic. It bounds the damage of every accepted model by the exact radius, and at the
published radius $\gamma=2$ that bound is loose, so an adversary that stays inside the radius can
drive the error to the floor. Section~\ref{sec:band} measures that case directly. Finally, the bounds extend to bounded-variance noise by using
$\mathbb E\|g_t-\nabla F(x_t)\|^2$ in place of $\zeta^2$, a standard step that we omit.

%% file: sections_jpdc/evaluation.tex
\section{Performance evaluation}
\label{sec:eval}

\subsection{Experimental setup}
\label{sec:eval-setup}

SketchGuard\footnote{Source code for Sketchguard is available at \url{https://github.com/murtazahr/Sketchguard}.} extends Murmura~\cite{murmura}, an open, configuration-driven DFL framework that
supplies the topologies, the attack suite, and the training loop. We add the screening interface,
the commitment and fetch message types on its ZeroMQ transport, and per-round byte accounting. We
train CNNs on three image datasets with 50 clients. EMNIST-balanced~\cite{cohen2017emnist} (47
classes, $d=833{,}967$) and CIFAR-10~\cite{krizhevsky2009} (10 classes, $d=1{,}070{,}794$) are split
by a Dirichlet partition ($\alpha_{\mathrm{dir}}=0.5$) with 300 and 500 training samples per client,
while FEMNIST~\cite{caldas2018leaf} (62 classes, $d=837{,}822$) keeps its natural partition of one
handwriting writer per client, so its heterogeneity comes from the writers rather than from a
synthetic split. Each client tests on 200 samples, and training uses SGD for 35 rounds. The large-model setting fine-tunes bert-mini~\cite{turc2019} (11.2M
parameters) on AG News~\cite{zhang2015agnews} across 10 clients for 12 rounds with AdamW. Training runs on an Erd\H{o}s--R\'enyi graph with $p=0.3$, with a density sweep in
Section~\ref{sec:sensitivity}. The continuum benchmark in Section~\ref{sec:cost} uses 20 nodes, 12 at the edge
(10\,Mbit/s up, 50\,Mbit/s down, 20\,ms delay) and 8 in the cloud (1\,Gbit/s, 2\,ms), on a sparse
random graph and on a cloud-hub topology in which every cloud node neighbors every node.

Reported \emph{Test error} (TER) is one minus test accuracy, averaged over honest clients and the last 3
rounds. \emph{Decision agreement} is the fraction of (receiver, neighbor, round) triples on which
the sketch screen and the exact test agree, \emph{pass rate} is the fraction of Byzantine attempts
the screen accepts, and a \emph{wasted fetch} is a Byzantine model fetched and then rejected. Bytes
follow the accounting of Section~\ref{sec:complexity}, and \emph{round time} is the wall-clock time
of the slowest honest node. \emph{SketchGuard} always wraps BALANCE unless stated; ablations are named by the rule they drop and
options by what they add. All runs use BALANCE's published defaults $\gamma=2$ and
$\kappa=1$~\cite{fang2024}, self-weight $\alpha=0.5$, and $k=400$, with attacks at 30\% Byzantine
unless stated, band placement $r=1.05$, and $n_c=16$ candidates. Every cell averages 3 seeds (2 for
105-round runs) on NVIDIA L40S GPUs, and runs are not bit-deterministic, which is one reason we
compare decisions rather than trajectories.

\subsection{Decision parity with the wrapped filters}
\label{sec:parity}

Compressed screening reproduces the decisions of both filters we wrap, at a fraction of the bytes.
Figure~\ref{fig:parity} pairs each filter with its SketchGuard wrapping on EMNIST, so the two bars
in a row should be the same length. Against BALANCE they agree under every attack, and across all
three datasets the largest gap in test error is \parityMaxGap, inside every seed standard deviation. This is what Lemma~\ref{lem:exact} predicts. Because SketchGuard aggregates
a subset of what the wrapped filter accepts, an attack BALANCE admits degrades both by the same
amount, and an attack BALANCE rejects is rejected by both.

\input{figures/fig_parity}

Panel (b) shows the other side of that fact. The saving is $\memnistgaussianBytes\times$ BALANCE for
the attacks the filter rejects and $\memnistalieBytes\times$ for ALIE and IPM, whose models the
filter accepts and would have fetched anyway. Screening therefore saves exactly what the filter
would have rejected, and nothing more. The median of five tables and 8-bit exchange move TER by at
most \txtRfiveDiff\ and \txtQeightCtsDiff.

The agreement is exact rather than approximate because of where these attacks sit. The screen
agrees with the exact test on \decAgreeHon\% of honest and Byzantine neighbors alike and loses
\decLost\% of honest neighbors, with honest agreement of \decfemnistAgreeHon\% on FEMNIST and
\deccifarOneZeroAgreeHon\% on CIFAR-10. None of the six standard attacks, however, places a model
inside the ambiguous band, and no honest pair enters it either, so this parity reflects attack
placement rather than a distortion-free sketch: measured distortion has standard deviation
\distStd\ against a theoretical $0.0707$, and at $\varepsilon=0.15$ Lemma~\ref{lem:cs} gives only
$\delta\approx0.2$ per pair.

Nothing in this depends on BALANCE. SCCLIP~\cite{he2022} clips every neighbor toward the receiver
rather than rejecting any, so as published it has no acceptance radius to wrap. Its clipping radius
is a natural one, and we therefore wrap \emph{screened SCCLIP}, the variant that accepts a neighbor
within that radius before clipping the accepted set. Lemma~\ref{lem:exact} applies to it unchanged,
since every aggregated model lies within the radius and clipping only shrinks its displacement
further, so the convergence results of Section~\ref{sec:convergence} carry over with no new
argument.

Figure~\ref{fig:parity} shows the same parity on both filters. Screening on sketches
rather than on exact distances changes its test error by at most \sccCompressGap\ across all
\sccCompressN\ attacks, while moving \sccBytesMin$\times$ the bytes wherever the filter rejects.
The acceptance rule of screened SCCLIP was not designed with sketching in mind, and the band-targeted
attack in particular is aimed at the sketch rather than at the filter, so the agreement is evidence
that compressed screening reproduces decisions rather than that the two rules happen to coincide.

\subsection{Effectiveness of the three rules}
\label{sec:rules}

\label{sec:band}
\input{sections/rev_tab_band}
Removing the exact test leaves the sketch to decide what is trusted, and a model placed in the
ambiguous band exploits that. At $r=1.05$ on EMNIST, the ablation reaches $\txtBandRawTER$ against
BALANCE's $\txtBandBalTER$, while SketchGuard stays at $\txtBandExTER$ (Table~\ref{tab:band}); the
sketch admits the model on \txtBandPass\% of attempts. The failure reproduces on the other datasets,
where the ablation reaches \mfemnistbandRaw\ against \mfemnistbandBal\ on FEMNIST and
\mcifarOneZerobandRaw\ against \mcifarOneZerobandBal\ on CIFAR-10, while SketchGuard matches BALANCE
(\mfemnistbandEx\ and \mcifarOneZerobandEx). Past the band, at $r=1.20$, the sketch passes no attempt
and all designs agree. Inside the radius, on the other hand, the wrapped filter accepts the model by
construction and so does every variant: at $r=0.80$ all three reach about \txtBandEightTER, the
Byzantine floor of Theorems~\ref{thm:pl} and~\ref{thm:nc}. At $r=0.95$ SketchGuard is slightly better
than BALANCE ($\txtBandNineFiveEx$ against $\txtBandNineFiveBal$), because the sketch rejects
\txtBandNineFiveRej\% of these in-radius models before they are fetched. The exact test therefore
settles whether the sketch adds an attack surface beyond the wrapped filter's own. It cannot make a
norm-relative filter robust to an adversary that stays inside the radius.

\input{figures/fig_amplification}
Removing the commitment lets the adversary build its model against a known map, and
Figure~\ref{fig:amplification} shows both outcomes of Theorem~\ref{thm:public} on EMNIST as the
Byzantine fraction grows. With a public seed and sketch-domain verification, TER rises to
$\txtAmpPubRawFourZero$ at 40\% Byzantine, the poisoning mode. Adding exact verification recovers
BALANCE's error but pins bytes at full exchange, the amplification mode. Commit-then-sketch with
exact verification recovers the same error with $\ampctsexOneZero$\%, $\ampctsexTwoZero$\%,
$\ampctsexThreeZero$\%, and $\ampctsexFourZero$\% fewer bytes at 10, 20, 30, and 40\% Byzantine,
tracking the rejection rate of \eqref{eq:ratio}. The same dichotomy holds at 30\% on the other
datasets, where the public seed with sketch-domain verification reaches \ampfemnistPubRawTER\ on
FEMNIST and \ampcifarOneZeroPubRawTER\ on CIFAR-10 and recovers \ampfemnistPubExTER\ and
\ampcifarOneZeroPubExTER\ with exact verification, at $\ampfemnistPubExBytes\times$ and
$\ampcifarOneZeroPubExBytes\times$ BALANCE's bytes. For bert-mini under the same attack at 40\%
Byzantine, the public-seed design with exact verification matches SketchGuard's error
($\txtLlmNullTER$) while moving \llmAmpRatio$\times$ its bytes.

Removing the ban lets a sketch be chosen after the seed. A forged sketch copied from the receiver's
own passes every screen, so each Byzantine edge forces a fetch every round. On EMNIST that is
\forgedCtSMismatch\ mismatched fetches per run, and the ablation moves $\forgedCtSRatio\times$
BALANCE's bytes, the amplification of Theorem~\ref{thm:public} regained. Test error is unaffected
(\forgedCtSTER), because the exact test rejects every forged model. With the ban, each of the
\forgedCtSbanBans\ Byzantine edges is caught at its first forged fetch, the run pays
\forgedCtSbanMismatch\ mismatched fetches, one per edge as Theorem~\ref{thm:ban} states, and bytes
return to $\forgedCtSbanRatio\times$ BALANCE. The same holds on FEMNIST
($\forgedfemnistNoBanBytes\times$ without the ban, $\forgedfemnistBanBytes\times$ with it) and
CIFAR-10 ($\forgedcifarOneZeroNoBanBytes\times$ and $\forgedcifarOneZeroBanBytes\times$). No honest
sender was banned in any run (\forgedCtSbanHonMismatch\ honest mismatches), and under directed
deviation, where Byzantine sketches are truthful, the ban never fires.

\label{sec:generalization}
\input{sections/rev_tab_gen}
Theorem~\ref{thm:public} holds for any linear map, so we repeat the attacks with screening on a
$k$-coordinate projection instead of a CS (Table~\ref{tab:gen}). With a fixed public coordinate set,
the null-space attack perturbs only unobserved coordinates: trusting the compressed test poisons the
aggregate (TER \gennullspacetopktopkbalancerawTER) and exact verification restores safety
(\gennullspacetopktopkbalanceexTER) while fetching every Byzantine model. Drawing the coordinates per
round after commitment removes both failures. The band-targeted attack shows the other half. Against
the seeded projection, whose distance estimate is far noisier than a CS's, the band model passes on
\genbandtopkctsrawPass\% of attempts and defeats compressed screening (TER
\genbandtopkctsrawTER), while the exact test holds TER at \genbandtopkctsexTER. A fixed projection,
conversely, passes no band model, because its estimate is biased toward the observed coordinates, and
it pays for that bias under the null-space attack. On this compressor, as on CS, seeding closes one
failure and exact verification the other, and neither suffices alone.

\subsection{Residual attacks}
\label{sec:residual-eval}
\input{sections/rev_tab_residual}
With all three rules in place, the residual attacks of Section~\ref{sec:residual} cost bandwidth
and nothing else, leaving test error untouched (Table~\ref{tab:residual}). Sparse cancellation passed the single-table sketch on
\scPassedRaw\ of \scAttempts\ attempts ($\txtScRate\%$), matching the $1/(2k)=1/800$ rate of
Theorem~\ref{thm:sparse}; raising the spike to $10^3\|\mu\|$ leaves TER at \scBigSpikeTER, and a
300-round run on 20 nodes reaches \scLongRawTER\ without the exact test and \scLongExTER\ with it.
Even unverified the damage is small, since a successful spike enters one receiver with weight
$(1-\alpha)/|\Sacc_i|$ and is diluted by later averaging. Select-after-commit turns extra candidates
into extra wasted fetches, from $\txtSacOne\%$ of attempts at $n_c=1$ to $\txtSacSixtyFour\%$ at
$n_c=64$, while TER stays between $\txtSacTERlo$ and $\txtSacTERhi$. The median of $R=5$ tables acts
on these wasted fetches, as Section~\ref{sec:residual} predicts. It removes sparse cancellation
(\scPassedRfive\ of \scAttempts\ attempts pass), cuts the band-targeted pass rate from
\resbandRone\% to \resbandRfive\%, and cuts select-after-commit from \resselectaftercommitRone\% to
\resselectaftercommitRfive\%. Five tables cost $0.19\%$ of a fetch against $0.05\%$ for one, and save
more than they cost: per-round bytes fall from \resBytesbandRone\,MB to \resBytesbandRfive\,MB under
the band-targeted attack and from \resBytesselectaftercommitRone\,MB to
\resBytesselectaftercommitRfive\,MB under select-after-commit, with TER unchanged.

\subsection{Cost on the edge--cloud continuum}
\label{sec:cost}

\label{sec:compute-eval}
\input{sections/rev_tab_scale}
Screening trades network cost for local compute, and Table~\ref{tab:scale} measures where that
trade pays. It reports one node's screening and aggregation step as the degree and the model grow.
The bytes received halve at every point, $\scaleBytesRatio\times$ BALANCE with half the neighbors
rejected, and the compute that buys this never exceeds \scaleShareMax\% of a round on a 50\,Mbit/s
uplink. Measured against BALANCE's compute alone the multiplier is larger,
$\scaleComputeMin\times$ to $\scaleComputeMax\times$, because BALANCE runs one distance test per
neighbor while SketchGuard adds a commitment opening and a re-sketch for every accepted neighbor and
pays a fixed seed expansion per round, both growing linearly in the number of neighbors and in $d$.
That multiplier, however, applies to a term the round barely notices. At $d=60$M and 299 neighbors,
SketchGuard spends $\scaleBigCtsT$\,s of compute against BALANCE's $\scaleBigBalT$\,s but receives
$\scaleBigCtsB$\,GB instead of $\scaleBigBalB$\,GB, so at 50\,Mbit/s the round falls from
$\scaleBigBalX$ to $\scaleBigCtsX$ minutes and compute accounts for \scaleBigCtsShare\% of it.
Equivalently, the break-even rate, the link speed below which the transfer saved outweighs the
compute added, lies between $\scaleBEMin$ and $\scaleBEMax$\,Gbit/s for every configuration, two to
three orders of magnitude above the edge uplinks below. These are per-node costs measured on one
node; no training run in this paper reaches 60M parameters.

Bytes behave the same way. With no attack, SketchGuard costs $\qCtSexzero\times$
BALANCE's bytes, since every neighbor is accepted and the sketch overhead is negligible. At 30\%
Byzantine under directed deviation it falls to $\qCtSexthirty\times$. Screening and quantization act
on different phases and therefore compose: BALANCE with 8-bit exchange costs
$\qBALqEightthirty\times$ BALANCE and SketchGuard with 8-bit exchange $\qCtSexqEightthirty\times$,
close to the product of the two savings.

\label{sec:cont-eval}
\input{sections/rev_tab_cont}
Table~\ref{tab:cont} measures bytes and wall-clock round time on real TCP with application-level link emulation, over 3 seeds, with
bytes as whole-run totals per round and timings over the last 3 of 4 rounds. The measured bytes
reproduce the accounting of Table~\ref{tab:bytes} exactly: every design that fetches every neighbor
receives $1.00\times$ BALANCE's bytes, including the amplified public-seed design, whose $0.05\%$
sketch overhead is below the table's resolution, while SketchGuard receives
$\contRelctsbThreeZeroB\times$ at 30\% Byzantine. Round time follows bytes on these links. The
amplified design takes $\contRelpublicbThreeZeroT\times$ BALANCE's round time, so amplification
erases the saving in wall clock as well as in bytes, whereas SketchGuard takes
$\contRelctsbThreeZeroT\times$, and $\contRelctsrelaybThreeZeroT\times$ with relay fetching. The gap
widens as edge uplinks tighten, reaching \contUpTwoBalT\,s for BALANCE against \contUpTwoCtsT\,s for
SketchGuard at 2\,Mbit/s.

Because relaying needs only the commitment, the
BALANCE-with-relay row separates the two mechanisms. On the sparse random graph, relaying alone gives
$\contRelbalancerelaybThreeZeroT\times$ BALANCE's round time at unchanged bytes, screening alone
$\contRelctsbThreeZeroT\times$, and both together $\contRelctsrelaybThreeZeroT\times$, so screening
carries most of the benefit. On the cloud-hub topology, however, where each edge model need be
uploaded only once, relaying alone reaches $\contHubRelbalancerelayT\times$ against
$\contHubRelctsT\times$ for screening alone and $\contHubRelctsrelayT\times$ together, so there most
of the saving comes from the relay rule. Every designated relay in the benchmark is honest, so
these round times exclude the timeout a withholding relay would add.

Without an attack, SketchGuard is slower than BALANCE
($\contRelctsbZeroT\times$), because the commit and sketch phases are barriers that direct exchange
does not have and no rejection offsets them. Relaying removes that gap
($\contRelctsrelaybZeroT\times$). Coupling the round to a beacon tick removes the barrier instead, at
the cost of dropping commitments that miss the tick: periods of 30\,s and 45\,s drop
\contBeaconThreeZeroDrop\% and \contBeaconFourFiveDrop\% of commitments, while a 20\,s period,
shorter than these links' transfer time, drops $\contBeaconTwoZeroDrop\%$.

\subsection{Sensitivity analysis}
\label{sec:sensitivity}

We vary three choices the analysis depends
on, a tighter radius ($\gamma\in\{0.5,1.0\}$), Metropolis mixing on the accepted subgraph
(Theorem~\ref{thm:topo}), and a $3\times$ longer budget. Honest decision agreement stays at or above
$\txtAgreeMin\%$ throughout, with TER matching BALANCE within noise, and after 105 rounds TER is
$\txtLongBal$ for BALANCE and $\txtLongCts$ for SketchGuard. Repeating the in-radius band attack
($r=0.95$), BALANCE falls from \txtGammaTwoBand\ at $\gamma=2$ to \txtGammaOneBand\ at $\gamma=1$
and \txtGammaHalfBand\ at $\gamma=0.5$, with SketchGuard at or slightly below it, as the floor term of
Theorems~\ref{thm:pl} and~\ref{thm:nc} predicts. The in-radius weakness belongs to the filter, not
to $\gamma=2$.

Theorem~\ref{thm:topo} predicts that error grows as the
honest graph mixes more slowly, through $1/(1-\lambda^2)$, and as data become more heterogeneous,
through $\zeta^2$. Both trends appear on EMNIST under directed deviation at 20\% Byzantine, and
SketchGuard follows BALANCE along each. Densifying the graph from $p=0.15$ to a complete graph lowers
TER from \topoSparseBal\ to \topoDenseBal\ for BALANCE and from \topoSparseSg\ to \topoDenseSg\ for
SketchGuard. Heterogeneity has the larger effect: moving from a near-IID split
($\alpha_{\mathrm{dir}}=10$) to a highly skewed one ($\alpha_{\mathrm{dir}}=0.05$) raises TER from
\hetIIDBal\ to \hetNonIIDBal\ and from \hetIIDSg\ to \hetNonIIDSg. FEMNIST adds a natural partition
across writers, and SketchGuard tracks BALANCE there as closely as on the Dirichlet splits.

%% file: figures/fig_parity.tex
% Non-floating on purpose: as a single-column float this figure was deferred to the end of the
% document by Overleaf's pdfLaTeX, and held back every later figure with it.
\begin{inlinefigure}
    \includegraphics[width=\columnwidth]{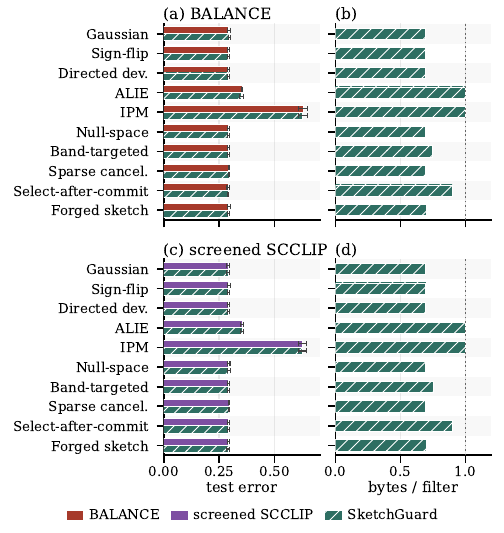}
    \caption{Decision parity and bytes for both wrapped filters (EMNIST, 30\% Byzantine, 3 seeds, $\pm$1 s.d.).}
    \label{fig:parity}
\end{inlinefigure}

%% file: sections/rev_tab_band.tex
\begin{table}[tbp]\centering\small
\caption{Band-targeted attack at varying distance from the acceptance radius.}
\label{tab:band}
\resizebox{\ifdim\width>\columnwidth \columnwidth\else\width\fi}{!}{%
\begin{tabular}{lcccc}\toprule
Distance / radius & BALANCE & w/o exact test & SketchGuard & Byz.\ pass sketch \\\midrule
0.80 & 0.909\,$\pm$\,0.005 & 0.909\,$\pm$\,0.003 & 0.910\,$\pm$\,0.005 & 100.0\% \\
0.95 & 0.901\,$\pm$\,0.008 & 0.852\,$\pm$\,0.068 & 0.859\,$\pm$\,0.018 & 73.7\% \\
1.05 & 0.292\,$\pm$\,0.006 & 0.682\,$\pm$\,0.088 & {\bfseries\boldmath 0.292\,$\pm$\,0.004} & 17.6\% \\
1.20 & 0.292\,$\pm$\,0.005 & 0.292\,$\pm$\,0.007 & 0.292\,$\pm$\,0.005 & 0.0\% \\
\bottomrule\end{tabular}}\end{table}

%% file: figures/fig_amplification.tex
\begin{figure*}[!tbp]
    \centering
    \subfloat[poisoning\label{fig:amp-ter}]{%
        \includegraphics[width=0.48\textwidth]{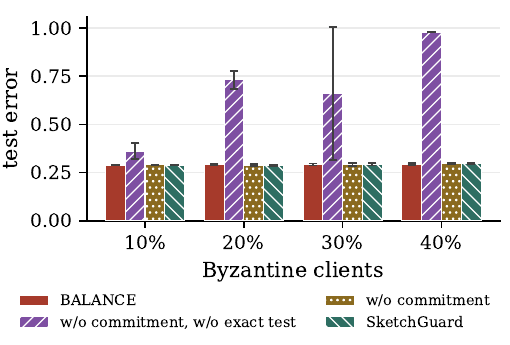}}
    \hfill
    \subfloat[bandwidth\label{fig:amp-bytes}]{%
        \includegraphics[width=0.48\textwidth]{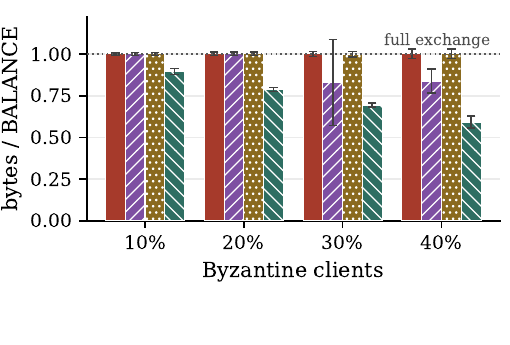}}
    \caption{Poisoning and amplification under the null-space attack.}
    \label{fig:amplification}
\end{figure*}

%% file: sections/rev_tab_gen.tex
\begin{table}[tbp]\centering\small
\caption{Adaptive attacks on a coordinate-projection compressor.}
\label{tab:gen}
\resizebox{\ifdim\width>\columnwidth \columnwidth\else\width\fi}{!}{%
\begin{tabular}{lllccc}\toprule
Attack & Coordinates & Verify & TER & Pass & Wasted/round \\\midrule
Null-space (top-$k$) & fixed & sketch & 0.940\,$\pm$\,0.060 & 94.0\% & -- \\
Null-space (top-$k$) & fixed & exact & 0.292\,$\pm$\,0.005 & 100.0\% & 159 \\
Null-space (top-$k$) & seeded & sketch & 0.292\,$\pm$\,0.005 & 0.0\% & -- \\
Null-space (top-$k$) & seeded & exact & {\bfseries\boldmath 0.292\,$\pm$\,0.005} & {\bfseries\boldmath 0.0\%} & {\bfseries\boldmath 0} \\
\midrule
Band-targeted & fixed & sketch & 0.293\,$\pm$\,0.005 & 0.0\% & -- \\
Band-targeted & fixed & exact & 0.293\,$\pm$\,0.005 & 0.0\% & 0 \\
Band-targeted & seeded & sketch & 0.858\,$\pm$\,0.048 & 53.4\% & -- \\
Band-targeted & seeded & exact & {\bfseries\boldmath 0.293\,$\pm$\,0.005} & {\bfseries\boldmath 46.9\%} & {\bfseries\boldmath 74} \\
\midrule
Select-after-commit & fixed & sketch & 0.292\,$\pm$\,0.005 & 55.7\% & -- \\
Select-after-commit & fixed & exact & 0.291\,$\pm$\,0.005 & 60.1\% & 96 \\
Select-after-commit & seeded & sketch & 0.292\,$\pm$\,0.005 & 15.0\% & -- \\
Select-after-commit & seeded & exact & {\bfseries\boldmath 0.290\,$\pm$\,0.005} & {\bfseries\boldmath 15.2\%} & {\bfseries\boldmath 24} \\
\bottomrule\end{tabular}}\end{table}

%% file: sections/rev_tab_residual.tex
\begin{table}[tbp]\centering\small
\caption{Residual attacks and the median-of-five estimator.}
\label{tab:residual}
\resizebox{\ifdim\width>\columnwidth \columnwidth\else\width\fi}{!}{%
\begin{tabular}{llcc}\toprule
Attack & Variant & TER & Byz.\ passed sketch \\\midrule
Sparse cancel. & w/o exact test & 0.294\,$\pm$\,0.002 & 20 of 16660 (0.12\%) \\
Sparse cancel. & SketchGuard & 0.294\,$\pm$\,0.002 & 20 of 16660 (0.12\%) \\
Sparse cancel. & SketchGuard, $R{=}5$ & 0.294\,$\pm$\,0.003 & 0 of 16660 (0.00\%) \\
\midrule
Band-targeted ($r=1.05$) & SketchGuard & 0.292\,$\pm$\,0.004 & 2926 of 16660 (17.56\%) \\
Band-targeted ($r=1.05$) & SketchGuard, $R{=}5$ & 0.293\,$\pm$\,0.005 & {\bfseries\boldmath 574 of 16660 (3.45\%)} \\
Select-after-commit ($n_c=16$) & SketchGuard & 0.291\,$\pm$\,0.006 & 11054 of 16660 (66.35\%) \\
Select-after-commit ($n_c=16$) & SketchGuard, $R{=}5$ & 0.291\,$\pm$\,0.005 & {\bfseries\boldmath 5637 of 16660 (33.84\%)} \\
\bottomrule\end{tabular}}\end{table}

%% file: sections/rev_tab_scale.tex
\begin{table}[tbp]\centering\small
\caption{Per-node compute and bytes against model size and degree.}
\label{tab:scale}
\resizebox{\ifdim\width>\columnwidth \columnwidth\else\width\fi}{!}{%
\begin{tabular}{rrcccccc}\toprule
 & & \multicolumn{2}{c}{Compute} & \multicolumn{2}{c}{Bytes received} & \\
$d$ & $N$ & BALANCE & SketchGuard & BALANCE & SketchGuard & Break-even \\\midrule
0.22M & 32 & 4\,ms & 21\,ms & 28\,MB & \textbf{14\,MB} & 6.7\,Gbit/s \\
0.83M & 16 & 10\,ms & 44\,ms & 53\,MB & \textbf{27\,MB} & 6.4\,Gbit/s \\
0.83M & 64 & 33\,ms & 157\,ms & 213\,MB & \textbf{107\,MB} & 6.9\,Gbit/s \\
0.83M & 299 & 148\,ms & 693\,ms & 997\,MB & \textbf{498\,MB} & 7.3\,Gbit/s \\
6.6M & 32 & 191\,ms & 895\,ms & 845\,MB & \textbf{422\,MB} & 4.8\,Gbit/s \\
11.2M & 32 & 459\,ms & 1.9\,s & 1.4\,GB & \textbf{718\,MB} & 3.9\,Gbit/s \\
60M & 32 & 2.9\,s & 10.4\,s & 7.7\,GB & \textbf{3.8\,GB} & 4.1\,Gbit/s \\
60M & 128 & 10.8\,s & 38.8\,s & 30.7\,GB & \textbf{15.4\,GB} & 4.4\,Gbit/s \\
60M & 299 & 24.9\,s & 88.7\,s & 71.8\,GB & \textbf{35.8\,GB} & 4.5\,Gbit/s \\
\bottomrule\end{tabular}}\end{table}

%% file: sections/rev_tab_cont.tex
\begin{table}[tbp]\centering\small
\caption{Round time and bytes at 30\% Byzantine on the emulated continuum.}
\label{tab:cont}
\resizebox{\ifdim\width>\columnwidth \columnwidth\else\width\fi}{!}{%
\begin{tabular}{llcc}\toprule
Topology & Design & honest rx bytes/round (MB) & round time (s) \\\midrule
sparse random & BALANCE & 292.4\,$\pm$\,31.5 & 28.68\,$\pm$\,4.06 \\
sparse random & BALANCE + relay & 292.5\,$\pm$\,31.5 & 27.18\,$\pm$\,6.48 \\
sparse random & w/o commitment (amplified) & 292.6\,$\pm$\,31.5 & 29.05\,$\pm$\,2.80 \\
sparse random & SketchGuard & 200.3\,$\pm$\,40.6 & 21.09\,$\pm$\,4.63 \\
sparse random & SketchGuard + relay & 200.3\,$\pm$\,40.6 & {\bfseries\boldmath 17.45\,$\pm$\,3.04} \\
\midrule
cloud-hub & BALANCE & 696.1\,$\pm$\,50.6 & 41.79\,$\pm$\,6.90 \\
cloud-hub & BALANCE + relay & 696.1\,$\pm$\,50.6 & 8.09\,$\pm$\,0.84 \\
cloud-hub & SketchGuard & 485.2\,$\pm$\,51.0 & 34.43\,$\pm$\,4.81 \\
cloud-hub & SketchGuard + relay & 485.2\,$\pm$\,51.0 & {\bfseries\boldmath 6.00\,$\pm$\,0.84} \\
\bottomrule\end{tabular}}\end{table}

%% file: sections_jpdc/discussion.tex
\section{Discussion}
\label{sec:discussion}

SketchGuard is not tied to a particular filter. It wraps any rule that accepts or rejects a neighbor
by Euclidean distance from the receiver's own model, the test at the heart of similarity-based
filtering, and Section~\ref{sec:parity} obtains the same parity on two such rules. BALANCE applies
the test directly, and a clipping aggregator such as SCCLIP adopts it by screening within the radius
it already clips to, gaining a bandwidth saving in the process. Rank-based rules such as
UBAR~\cite{guo2022} and radius-free rules such as geometric-median
aggregation~\cite{pillutla2022} lie outside the guarantee as stated, and extending it to them is a
natural next step. The front-end is equally indifferent to how it compresses.
Theorem~\ref{thm:public} threatens every linear compressor into fewer than $d$ dimensions, and
Table~\ref{tab:gen} reproduces both failures and both repairs on a coordinate projection, so the
compressor sets the bandwidth and never the safety. With all three rules in place the sketch adds no
poisoning surface, and the residual attacks of Section~\ref{sec:residual} cost only bounded wasted
fetches. The adversary's only remaining lever is a model inside the acceptance radius, which it
already holds against the uncompressed filter, and closing that gap needs a better filter rather
than a better sketch.

Two choices then shape a deployment, the seed source and the round clock. A cloud-hosted beacon in
the style of drand~\cite{drand2020} is the simplest source, since every client reads the same value
and trusts no other client's randomness, but the round then depends on that beacon staying available
and unpredictable until the commitment barrier closes, whereas a commit-reveal of nonces among the
peers removes the dependency at the cost of one more broadcast per round and at least one honest
contributor. Coupling the round to the beacon tick removes the cost of the commit and sketch
barriers, but under intermittent connectivity the commitments that miss a tick thin the honest
subgraph on which Theorem~\ref{thm:topo} depends. The beacon period should therefore be set from the
measured fetch time of the slowest link class, with the per-round drop rate as the quantity to
monitor. Because commitments, seeds, and sketches are fresh every round, dynamic topologies need no
extra coordination.

%% file: sections_jpdc/conclusion.tex
\section{Conclusions and Future Directions}
\label{sec:conclusion}

We proposed SketchGuard, which breaks the coupling between Byzantine filtering and full-model
communication in DFL. Clients screen neighbors on compact sketches and fetch full models only from
those that pass, cutting per-round communication from $O(d|\N_i|)$ to $O(k|\N_i|+d|\Sacc_i|)$ with a
sketch size independent of the model dimension.

Screening on a summary, however, hands the adversary whatever that summary hides. Knowing the map
lets it hide a perturbation of any size from the sketch, which either poisons the aggregate or forces
a full transfer on every Byzantine edge, and knowing the seed lets it forge a sketch that passes the
screen. Three rules close these gaps, and none substitutes for another. Exact post-fetch
verification makes the aggregated set a subset of what the wrapped filter accepts, so safety never
rests on the sketch. Committing to a model before the seed is drawn keeps the bandwidth saving under
attack. Banning a sender whose model does not match its sketch caps a forged sketch at one wasted
fetch per edge. The wrapped filter's convergence guarantees carry over unchanged.

Across three datasets, two wrapped filters, and two compressors, SketchGuard reproduces the
decisions of the filter it wraps to within \parityMaxGap\ test error, while a sketch-only screen is
defeated by a model placed in the ambiguous band (\memnistbandRaw\ against \memnistbandEx). Under
attack it moves $\memnistgaussianBytes\times$ the bytes of full exchange for under \scaleShareMax\%
of a round in compute, and with relaying it cuts the round on emulated edge and cloud links to
$\contRelctsrelaybThreeZeroT\times$ that of BALANCE.

Future work will relax the bounded-iterate assumption and carry the analysis to stochastic
gradients, move the benchmark to a geo-distributed testbed at larger scale, and extend the
commitment principle to asynchronous and churn-heavy deployments.

%% file: sections_jpdc/appendix.tex
\appendix
\renewcommand{\thesection}{Appendix~\Alph{section}}
\renewcommand{\theequation}{\Alph{section}.\arabic{equation}}
\makeatletter\@addtoreset{equation}{section}\makeatother
\setcounter{equation}{0}
\allowdisplaybreaks
% The appendix is display-dense: default display skips dominate its height.
\setlength{\abovedisplayskip}{4pt plus 1pt minus 1pt}
\setlength{\belowdisplayskip}{4pt plus 1pt minus 1pt}
\setlength{\abovedisplayshortskip}{2pt plus 1pt}
\setlength{\belowdisplayshortskip}{2pt plus 1pt}

\section{Proofs for Section~\ref{sec:design}}
\label{app:security}

Throughout, $c_t=\gamma e^{-\kappa t/T}$ is the round's decayed multiplier, so the exact radius is
$\tau_i^t=c_t\|\bw_i\|$ and the sketch-domain radius is $\hat\tau_i^t=c_t\|\bs_i\|$. We write
$\bw_i$ for $\bw_i^{t+1/2}$ and take $0\le\varepsilon<1$.

\begin{proof}[Proof of Lemma~\ref{lem:screen}]
Taking square roots in the two conditions of $E_\varepsilon(i,j)$ gives
\begin{align}
\sqrt{1-\varepsilon}\,D_{ij}&\le\hat d_{ij}\le\sqrt{1+\varepsilon}\,D_{ij},\label{eq:app-dist}\\
\sqrt{1-\varepsilon}\,\|\bw_i\|&\le\|\bs_i\|\le\sqrt{1+\varepsilon}\,\|\bw_i\|.\label{eq:app-norm}
\end{align}

\noindent \emph{Part (a).} If the screen accepts $j$, then $\hat d_{ij}\le c_t\|\bs_i\|$. The left side of
\eqref{eq:app-dist} and the right side of \eqref{eq:app-norm} give
$\sqrt{1-\varepsilon}\,D_{ij}\le c_t\sqrt{1+\varepsilon}\,\|\bw_i\|$, so
\begin{equation}
D_{ij}\le c_t\sqrt{\tfrac{1+\varepsilon}{1-\varepsilon}}\,\|\bw_i\|
=\geff e^{-\kappa t/T}\|\bw_i\| .
\end{equation}

\noindent \emph{Part (b).} Suppose $D_{ij}\le\gin e^{-\kappa t/T}\|\bw_i\|$, that is,
$D_{ij}\le c_t\sqrt{(1-\varepsilon)/(1+\varepsilon)}\,\|\bw_i\|$. The right side of
\eqref{eq:app-dist} and the left side of \eqref{eq:app-norm} give
\begin{align}
\hat d_{ij}&\le\sqrt{1+\varepsilon}\,D_{ij}\notag\\
&\le c_t\sqrt{1-\varepsilon}\,\|\bw_i\|\le c_t\|\bs_i\| ,
\end{align}
so the screen accepts $j$.
\end{proof}

\begin{proof}[Proof of Lemma~\ref{lem:median}]
Let $x_1,\dots,x_R$ be the per-table estimates and $I=[(1-\varepsilon)D^2,(1+\varepsilon)D^2]$.
More than $R/2$ estimates lie in $I$, so fewer than $R/2$ lie below it and fewer than $R/2$ lie
above it. Any median $m$ has at least $R/2$ estimates at or below it and at least $R/2$ at or above
it. If $m$ were below $I$, all the estimates at or below $m$ would also lie below $I$, which is
impossible. The case of $m$ above $I$ is symmetric, so $m\in I$.

For the tail bound, let $Y_r\in\{0,1\}$ indicate that table $r$ fails, meaning $x_r\notin I$. The
tables use independent hash and sign functions, so the $Y_r$ are independent with means at most
$p$. By the first part, the median fails only if $\sum_rY_r\ge R/2$. Hoeffding's
inequality~\cite{hoeffding1963} for independent variables in $[0,1]$ gives
\begin{align}
&\Pr\Big[\sum_rY_r\ge\tfrac R2\Big]\notag\\
&\quad\le\Pr\Big[\sum_rY_r-\mathbb E\sum_rY_r\ge R\big(\tfrac12-p\big)\Big]\notag\\
&\quad\le\exp\!\big(-2R(\tfrac12-p)^2\big). \qedhere
\end{align}
\end{proof}

\begin{proof}[Proof of Theorem~\ref{thm:cost}]
\emph{(i)} A wasted fetch means the screen accepts $j$ and the exact test rejects it. Rejection
gives $D_{ij}>\tau_i^t$, and acceptance gives $D_{ij}\le\geff e^{-\kappa t/T}\|\bw_i\|$ by
Lemma~\ref{lem:screen}(a).

For (ii) and (iii), note first that $\bw_i$ (committed by the honest client) and $\bw_j$
(seed-oblivious by hypothesis) are both fixed independently of $\theta_t$. Lemma~\ref{lem:cs},
applied to the pair $(\bw_i,\bw_j)$ and to the pair $(\bw_i,\mathbf0)$, shows that each of the two
estimates in $E_\varepsilon(i,j)$ is inaccurate with probability at most $\delta$. A union bound
gives
\begin{equation}
\Pr[E_\varepsilon(i,j)\text{ fails}]\le2\delta .
\end{equation}

\noindent \emph{(ii)} If $D_{ij}>\geff e^{-\kappa t/T}\|\bw_i\|$ and the sketch is accurate, the
contrapositive of Lemma~\ref{lem:screen}(a) shows that the screen rejects $j$. Such a model is
therefore fetched only if the sketch is inaccurate, which has probability at most $2\delta$.

\noindent \emph{(iii)} If $D_{ij}\le\gin e^{-\kappa t/T}\|\bw_i\|$ and the sketch is accurate,
Lemma~\ref{lem:screen}(b) shows that the screen accepts $j$. Such a model is therefore skipped only if
the sketch is inaccurate, which has probability at most $2\delta$.
\end{proof}

\begin{proof}[Proof of Theorem~\ref{thm:public}]
The map $A$ has rank at most $k$. By the rank--nullity theorem, its kernel has dimension
\begin{equation}
\dim\ker A=d-\operatorname{rank}A\ge d-k\ge1 .
\end{equation}
Choose a nonzero $\bv\in\ker A$ and any $M>\tau$, and set $\tilde\bw=\bw+M\bv/\|\bv\|$. Then
\begin{equation}
A\tilde\bw=A\bw,\qquad\|\tilde\bw-\bw\|=M>\tau .
\end{equation}
The sender reports the true sketch $A\tilde\bw$, whose distance to $A\bw$ is zero. It therefore
passes every sketch screen against the reference $\bw$ and the re-sketch check after the fetch,
and it fails the exact test. Without the exact test, it enters the aggregate with a positive weight
and moves it by an amount that grows with $M$, which is arbitrary. With the exact test, it is
fetched and then rejected, so each such neighbor forces one wasted fetch in every round.
\end{proof}

\begin{proof}[Proof of Theorem~\ref{thm:cts}]
By Assumption~\ref{as:commit}, a delivered model that differs from the committed one fails to open
its commitment, except with negligible probability, and is dropped in Phase~\ref{ph:verify}. Every
fetched model is therefore the one its sender committed in Phase~\ref{ph:commit}, which depends
only on information available before the commitment barrier. The seed is drawn after that barrier
and independently of everything committed before it, so every delivered model is seed-oblivious.

Conditional on the committed models, the pair $(\bw_i,\bw_j)$ is fixed and $\theta_t$ is fresh, so
Lemma~\ref{lem:cs} applies to it as to any fixed pair. If moreover $\bs_j=\CS_{\theta_t}(\bw_j)$,
the screen decision for $j$ is the one analyzed in Theorem~\ref{thm:cost}. That theorem bounds the
probability of a wasted fetch by $2\delta$ for a model outside the band. A model inside the band
wastes at most one fetch per round, since each neighbor is fetched at most once per round. Safety
follows from Lemma~\ref{lem:exact}, which does not involve the sketch.
\end{proof}

\begin{proof}[Proof of Theorem~\ref{thm:ban}]
Channels are reliable, and an honest sender computes its sketch from the model it commits, then
delivers that model with its nonce. The delivered pair opens the commitment and its re-sketch
equals the reported sketch, so an honest sender is never banned.

Now fix one edge, direct or through a relay. Call a round \emph{wasted} on that edge if a delivery
over it fails to open, or opens but does not match its reported sketch. Either outcome proves
misbehavior by the peer that delivered it, and that peer is banned, so no delivery crosses the edge
after the first wasted round. Two wasted rounds $t_1<t_2$ would require a delivery after the ban
imposed at $t_1$, which is impossible. Each edge therefore has at most one wasted round, and summing
over any set $E_B$ of edges gives at most $|E_B|$.
\end{proof}

\begin{proof}[Proof of Theorem~\ref{thm:sparse}]
Let $\mathbf u_b$ be the $b$-th standard basis vector of $\R^k$. By linearity,
\begin{align}
&A\big(\mu+c(\mathbf e_a-\mathbf e_b)\big)\notag\\
&\qquad=A\mu+c\big(\sigma(a)\,\mathbf u_{h(a)}-\sigma(b)\,\mathbf u_{h(b)}\big).
\end{align}
There are three cases.
\begin{itemize}
\item If $h(a)=h(b)$ and $\sigma(a)=\sigma(b)$, the two terms cancel, and the spiked model has the
sketch of $\mu$.
\item If $h(a)=h(b)$ and $\sigma(a)\ne\sigma(b)$, bucket $h(a)$ moves by $2|c|$.
\item If $h(a)\ne h(b)$, bucket $h(a)$ moves by $|c|$.
\end{itemize}
The tuple $(h(a),h(b),\sigma(a),\sigma(b))$ takes $k\cdot k\cdot2\cdot2=4k^2$ equally likely values.
Exactly $2k$ of them cancel, namely a common bucket ($k$ choices) with a common sign (two choices).
The cancellation probability is therefore $2k/(4k^2)=1/(2k)$, whatever the seed.
\end{proof}

\paragraph{Sparse cancellation under the median.}
With $R$ independent tables, a spike of size $|c|$ moves some bucket of every non-cancelling table
by at least $|c|$. Once $|c|$ exceeds the honest spread, that table's distance estimate is large.
The median estimate therefore stays small only if at least $\lceil R/2\rceil$ tables cancel at
once. Each table cancels independently with probability $1/(2k)$, so a union bound over the
$\binom{R}{\lceil R/2\rceil}$ subsets of that size bounds the probability by
\begin{equation}
\binom{R}{\lceil R/2\rceil}(2k)^{-\lceil R/2\rceil}.
\end{equation}

\section{Proofs for Section~\ref{sec:convergence}}
\label{app:convergence}

Write $\nabla F$ for the gradient of $F$ and $C=\tfrac{3\eta}4\zeta^2+\rho\beta+L\beta^2$ for the
per-round error term.

\begin{proof}[Proof of Theorem~\ref{thm:onestep}]
Let $\bu=-\eta g_t+\Delta$, so the new point is $y=x+\bu$. Write $g_t=\nabla F(x)+\mathbf e$ with
$\|\mathbf e\|\le\zeta$. By $L$-smoothness,
\begin{equation}\label{eq:app-smooth}
F(y)\le F(x)+\langle\nabla F(x),\bu\rangle+\tfrac L2\|\bu\|^2 .
\end{equation}

\noindent \emph{Inner product.} Expanding $\bu$ gives
\begin{align}
\langle\nabla F(x),\bu\rangle
&=-\eta\|\nabla F(x)\|^2-\eta\langle\nabla F(x),\mathbf e\rangle\notag\\
&\quad+\langle\nabla F(x),\Delta\rangle .
\end{align}
Cauchy--Schwarz with $\|\nabla F(x)\|\le\rho$ and $\|\Delta\|\le\beta$ bounds the last term by
$\rho\beta$. For the middle term, Cauchy--Schwarz and then Young's inequality give
$|\langle\nabla F(x),\mathbf e\rangle|\le\tfrac12\|\nabla F(x)\|^2+\tfrac12\zeta^2$. Hence
\begin{equation}\label{eq:app-inner}
\langle\nabla F(x),\bu\rangle\le-\tfrac\eta2\|\nabla F(x)\|^2+\tfrac\eta2\zeta^2+\rho\beta .
\end{equation}

\noindent \emph{Quadratic term.} Applying $\|\mathbf a+\mathbf b\|^2\le2\|\mathbf a\|^2+2\|\mathbf b\|^2$
twice gives $\|\bu\|^2\le2\eta^2\|g_t\|^2+2\beta^2$ and $\|g_t\|^2\le2\|\nabla F(x)\|^2+2\zeta^2$.
Therefore
\begin{align}
\tfrac L2\|\bu\|^2
&\le2L\eta^2\|\nabla F(x)\|^2+2L\eta^2\zeta^2+L\beta^2\notag\\
&\le\tfrac\eta4\|\nabla F(x)\|^2+\tfrac\eta4\zeta^2+L\beta^2 ,\label{eq:app-quad}
\end{align}
where the last step uses $2L\eta^2\le\eta/4$, which is $\eta L\le1/8$.

\noindent \emph{Conclusion.} Substituting \eqref{eq:app-inner} and \eqref{eq:app-quad} into
\eqref{eq:app-smooth} gives
\begin{equation}
F(y)\le F(x)-\tfrac\eta4\|\nabla F(x)\|^2+C . \qedhere
\end{equation}
\end{proof}

\begin{proof}[Proof of Theorem~\ref{thm:pl}]
Let $e_t=F(x_t)-F^\star$ and $q=1-\mu\eta/2$. Theorem~\ref{thm:onestep} and the PL inequality
$\|\nabla F(x_t)\|^2\ge2\mu e_t$ give $e_{t+1}\le q\,e_t+C$, and $0\le q<1$ because $\mu\le L$ and
$\eta L\le1/8$ give $\mu\eta/2\le1/16$. Unrolling the recursion by induction on $T$ gives
$e_T\le q^Te_0+C\sum_{s<T}q^s$, and bounding the geometric sum by $1/(1-q)=2/(\mu\eta)$ gives
$e_T\le q^Te_0+2C/(\mu\eta)$. Substituting $C$, the floor is
$2C/(\mu\eta)=3\zeta^2/(2\mu)+2(\rho\beta+L\beta^2)/(\mu\eta)$.
\end{proof}

\begin{proof}[Proof of Theorem~\ref{thm:nc}]
Rearranging Theorem~\ref{thm:onestep} at round $t$ gives
$\tfrac\eta4\|\nabla F(x_t)\|^2\le F(x_t)-F(x_{t+1})+C$. Summing over $t<T$, the right-hand side
telescopes, and since $F(x_T)\ge F^\star$ this gives
$\tfrac\eta4\sum_{t<T}\|\nabla F(x_t)\|^2\le F(x_0)-F^\star+TC$. Dividing by $\eta T/4$ and using
$4C/\eta=3\zeta^2+4(\rho\beta+L\beta^2)/\eta$ gives the statement.
\end{proof}

\begin{proof}[Proof of Theorem~\ref{thm:topo}]
Let $x_t=\bar\bw^t$. Since $\nabla F(\bar\bw^t)=\frac1{|H|}\sum_{i\in H}\nabla f_i(\bar\bw^t)$,
the gradient error is
\begin{equation}
g_t-\nabla F(\bar\bw^t)
=\frac1{|H|}\sum_{i\in H}\big(\nabla f_i(\bw_i^t)-\nabla f_i(\bar\bw^t)\big).
\end{equation}
Each honest $f_i$ is $L$-smooth (Assumption~\ref{as:sc}), so each summand has norm at most
$L\|\bw_i^t-\bar\bw^t\|$. Jensen's inequality for the squared norm then gives
\begin{equation}
\zeta_t^2:=\|g_t-\nabla F(\bar\bw^t)\|^2\le L^2\,\Omega_t .
\end{equation}
Theorem~\ref{thm:onestep} holds in every round with its own error $\zeta_t$. The telescoping
argument of Theorem~\ref{thm:nc}, with $\zeta^2$ replaced by $\zeta_t^2$ in round $t$, gives
\begin{align}
\frac1T\sum_{t<T}\|\nabla F(\bar\bw^t)\|^2
&\le\frac{4(F(\bar\bw^0)-F^\star)}{\eta T}\notag\\
&\quad+\frac{3L^2}T\sum_{t<T}\Omega_t+\frac{4(\rho\beta+L\beta^2)}\eta .\label{eq:app-topo}
\end{align}
Unrolling the consensus recursion as in the proof of Theorem~\ref{thm:pl}, now with ratio
$\lambda^2<1$, gives $\Omega_t\le\lambda^{2t}\Omega_0+B_\xi/(1-\lambda^2)$. Averaging over $t<T$
and using $\sum_{t<T}\lambda^{2t}\le1/(1-\lambda^2)$,
\begin{equation}
\frac1T\sum_{t<T}\Omega_t\le\frac{\Omega_0}{T(1-\lambda^2)}+\frac{B_\xi}{1-\lambda^2}.
\end{equation}
Substituting this bound into \eqref{eq:app-topo} yields the statement. Every step is linear in
$\Omega_t$ and in the squared gradient norms, so the same argument applied to their expectations
gives the bound in expectation whenever the consensus recursion holds in expectation, as it does
with the coefficients of~\cite{koloskova2020}.
\end{proof}

%% file: refs.bib
@inproceedings{lian2017,
  title={Can decentralized algorithms outperform centralized algorithms? A case study for decentralized parallel stochastic gradient descent},
  author={Lian, Xiangru and Zhang, Ce and others},
  booktitle={Advances in Neural Information Processing Systems}, year={2017}}

@article{martinez2023,
  title={Decentralized federated learning: Fundamentals, state of the art, frameworks, trends, and challenges},
  author={Mart{\'\i}nez Beltr{\'a}n, Enrique Tom{\'a}s and Quiles P{\'e}rez, Mario and others},
  journal={IEEE Communications Surveys \& Tutorials}, pages={2983--3013}, year={2023}}

@inproceedings{fang2024,
  title={{Byzantine-robust} decentralized federated learning},
  author={Fang, Minghong and Zhang, Zifan and others},
  booktitle={Proceedings of ACM SIGSAC Conference on Computer and Communications Security}, pages={2874--2888}, year={2024}}

@article{he2022,
  title={{Byzantine-robust} decentralized learning via {ClippedGossip}},
  author={He, Lie and Karimireddy, Sai Praneeth and Jaggi, Martin},
  journal={arXiv preprint arXiv:2202.01545}, year={2022}}

@article{guo2022,
  title={{Byzantine-resilient} decentralized stochastic gradient descent},
  author={Guo, Shangwei and Zhang, Tianwei and others},
  journal={IEEE Transactions on Circuits and Systems for Video Technology}, volume={32}, number={6}, pages={4096--4106}, year={2022}}

@article{cajaraville2025,
  title={{Byzantine-robust} aggregation for securing decentralized federated learning},
  author={Cajaraville-Aboy, Diego and Fern{\'a}ndez-Vilas, Ana and others},
  journal={IEEE Access}, volume={13}, pages={190947--190963}, year={2025},
  doi={10.1109/ACCESS.2025.3629864}}

@article{ams1999,
  title={The space complexity of approximating the frequency moments},
  author={Alon, Noga and Matias, Yossi and Szegedy, Mario},
  journal={Journal of Computer and System Sciences}, volume={58}, number={1}, pages={137--147}, year={1999},
  doi={10.1006/jcss.1997.1545}}

@article{thorup2012,
  title={Tabulation-based 5-independent hashing with applications to linear probing and second moment estimation},
  author={Thorup, Mikkel and Zhang, Yin},
  journal={SIAM Journal on Computing}, volume={41}, number={2}, pages={293--331}, year={2012},
  doi={10.1137/100800774}}

@inproceedings{cohen2017emnist,
  title={{EMNIST}: Extending {MNIST} to handwritten letters},
  author={Cohen, Gregory and Afshar, Saeed and others},
  booktitle={Proceedings of the International Joint Conference on Neural Networks}, pages={2921--2926}, year={2017}}

@techreport{krizhevsky2009,
  title={Learning multiple layers of features from tiny images},
  author={Krizhevsky, Alex},
  institution={University of Toronto}, year={2009}}

@article{caldas2018leaf,
  title={{LEAF}: A benchmark for federated settings},
  author={Caldas, Sebastian and Duddu, Sai Meher Karthik and others},
  journal={arXiv preprint arXiv:1812.01097}, year={2018}}

@article{turc2019,
  title={Well-read students learn better: On the importance of pre-training compact models},
  author={Turc, Iulia and Chang, Ming-Wei and others},
  journal={arXiv preprint arXiv:1908.08962}, year={2019}}

@inproceedings{zhang2015agnews,
  title={Character-level convolutional networks for text classification},
  author={Zhang, Xiang and Zhao, Junbo and LeCun, Yann},
  booktitle={Advances in Neural Information Processing Systems}, volume={28}, pages={649--657}, year={2015}}

@inproceedings{charikar2002,
  title={Finding frequent items in data streams},
  author={Charikar, Moses and Chen, Kevin and Farach-Colton, Martin},
  booktitle={Proceedings of the International Colloquium on Automata, Languages, and Programming}, pages={693--703}, year={2002}}

@inproceedings{hardt2013,
  title={How robust are linear sketches to adaptive inputs?},
  author={Hardt, Moritz and Woodruff, David P},
  booktitle={Proceedings of the ACM Symposium on Theory of Computing}, pages={121--130}, year={2013}}

@inproceedings{cohen2022,
  title={On the robustness of {CountSketch} to adaptive inputs},
  author={Cohen, Edith and Lyu, Xin and others},
  booktitle={Proceedings of the International Conference on Machine Learning}, year={2022}}

@article{beneliezer2022,
  title={A framework for adversarially robust streaming algorithms},
  author={Ben-Eliezer, Omri and Jayaram, Rajesh and others},
  journal={Journal of the ACM}, volume={69}, number={2}, year={2022}}

@inproceedings{gorbunov2022,
  title={Variance reduction is an antidote to {Byzantines}: Better rates, weaker assumptions and communication compression as a cherry on the top},
  author={Gorbunov, Eduard and Horv{\'a}th, Samuel and others},
  booktitle={Proceedings of the International Conference on Learning Representations}, year={2023}}

@inproceedings{rammal2024,
  title={Communication compression for {Byzantine} robust learning: New efficient algorithms and improved rates},
  author={Rammal, Ahmad and Gruntkowska, Kaja and others},
  booktitle={Proceedings of the International Conference on Artificial Intelligence and Statistics}, pages={1207--1215}, year={2024}}

@inproceedings{xia2025,
  title={{Fed-DPRoC}: Communication-efficient differentially private and robust federated learning},
  author={Xia, Yue and Jahani-Nezhad, Tayyebeh and Bitar, Rawad},
  booktitle={Proceedings of the International Conference on Federated Learning Technologies and Applications}, pages={142--149}, year={2025}}

@inproceedings{rothchild2020,
  title={{FetchSGD}: Communication-efficient federated learning with sketching},
  author={Rothchild, Daniel and Panda, Ashwinee and others},
  booktitle={Proceedings of the International Conference on Machine Learning}, year={2020}}

@inproceedings{baruch2019,
  title={A little is enough: Circumventing defenses for distributed learning},
  author={Baruch, Gilad and Baruch, Moran and Goldberg, Yoav},
  booktitle={Advances in Neural Information Processing Systems}, year={2019}}

@inproceedings{xie2020,
  title={Fall of empires: Breaking {Byzantine-tolerant} {SGD} by inner product manipulation},
  author={Xie, Cong and Koyejo, Oluwasanmi and Gupta, Indranil},
  booktitle={Proceedings of Conference on Uncertainty in Artificial Intelligence}, pages={261--270}, year={2020}}

@inproceedings{koloskova2020,
  title={A unified theory of decentralized {SGD} with changing topology and local updates},
  author={Koloskova, Anastasia and Loizou, Nicolas and others},
  booktitle={Proceedings of the International Conference on Machine Learning}, year={2020}}

@inproceedings{drand2020,
  title={Scalable bias-resistant distributed randomness},
  author={Syta, Ewa and Jovanovic, Philipp and others},
  booktitle={Proceedings of the IEEE Symposium on Security and Privacy (S\&P)}, pages={444--460}, year={2017}}

@inproceedings{murmura,
  author={Rangwala, Murtaza and Sinnott, Richard O. and Buyya, Rajkumar},
  booktitle={Proceedings of the IEEE 26th International Symposium on Cluster, Cloud and Internet Computing},
  title={Evidential Trust-Aware Model Personalization in Decentralized Federated Learning for Wearable {IoT}},
  year={2026},
  pages={510--519}}

@inproceedings{signsgd2018,
  title={{signSGD}: Compressed optimisation for non-convex problems},
  author={Bernstein, Jeremy and Wang, Yu-Xiang and others},
  booktitle={Proceedings of the International Conference on Machine Learning}, pages={560--569}, year={2018}}

@inproceedings{qsgd2017,
  title={{QSGD}: Communication-efficient {SGD} via gradient quantization and encoding},
  author={Alistarh, Dan and Grubic, Demjan and others},
  booktitle={Advances in Neural Information Processing Systems}, year={2017}}

@inproceedings{koloskova2019,
  title={Decentralized stochastic optimization and gossip algorithms with compressed communication},
  author={Koloskova, Anastasia and Stich, Sebastian U and Jaggi, Martin},
  booktitle={Proceedings of the International Conference on Machine Learning}, pages={3478--3487}, year={2019}}

@article{kairouz2021,
  title={Advances and open problems in federated learning},
  author={Kairouz, Peter and McMahan, H Brendan and others},
  journal={Foundations and Trends in Machine Learning}, volume={14}, number={1--2}, pages={1--210}, year={2021}}

@article{lim2020,
  title={Federated learning in mobile edge networks: A comprehensive survey},
  author={Lim, Wei Yang Bryan and Luong, Nguyen Cong and others},
  journal={IEEE Communications Surveys \& Tutorials}, volume={22}, number={3}, pages={2031--2063}, year={2020}}

@inproceedings{scaffold2020,
  title={{SCAFFOLD}: Stochastic controlled averaging for federated learning},
  author={Karimireddy, Sai Praneeth and Kale, Satyen and others},
  booktitle={Proceedings of the International Conference on Machine Learning}, pages={5132--5143}, year={2020}}

@inproceedings{krum2017,
  title={Machine learning with adversaries: {Byzantine} tolerant gradient descent},
  author={Blanchard, Peva and El Mhamdi, El Mahdi and others},
  booktitle={Advances in Neural Information Processing Systems}, volume={30}, year={2017}}

@article{rosendo2022,
  title={Distributed intelligence on the {Edge-to-Cloud} Continuum: A systematic literature review},
  author={Rosendo, Daniel and Costan, Alexandru and others},
  journal={Journal of Parallel and Distributed Computing}, volume={166}, pages={71--94}, year={2022}}

@article{hegedus2021,
  title={Decentralized learning works: An empirical comparison of gossip learning and federated learning},
  author={Heged{\H{u}}s, Istv{\'a}n and Danner, G{\'a}bor and Jelasity, M{\'a}rk},
  journal={Journal of Parallel and Distributed Computing}, volume={148}, pages={109--124}, year={2021},
  doi={10.1016/j.jpdc.2020.10.006}}

@article{dai2024sketchfusion,
  title={{Sketch-fusion}: A gradient compression method with multi-layer fusion for communication-efficient distributed training},
  author={Dai, Li-Rong and Gong, Luqi and others},
  journal={Journal of Parallel and Distributed Computing}, volume={185}, pages={104811}, year={2024}}

@article{li2025brfl,
  title={{BRFL}: A blockchain-based byzantine-robust federated learning model},
  author={Li, Yang and Xia, Chunhe and others},
  journal={Journal of Parallel and Distributed Computing}, volume={196}, pages={104995}, year={2025}}

@article{zeng2025poisonproof,
  title={Towards value-sensitive and poisoning-proof model aggregation for federated learning on heterogeneous data},
  author={Zeng, Hui and Zhou, Tongqing and others},
  journal={Journal of Parallel and Distributed Computing}, volume={196}, year={2025}}

@article{wang2025b2dfl,
  title={{B2DFL}: Bringing butterfly to decentralized federated learning assisted with blockchain},
  author={Wang, Hao and Cai, Yichen and others},
  journal={Journal of Parallel and Distributed Computing}, volume={195}, pages={104978}, year={2025}}

@article{han2025owl,
  title={{OWL}: Worker-assisted server bandwidth optimization for efficient communication federated learning},
  author={Han, Xiaoming and Liu, Boan and Hu, Chuang and Cheng, Dazhao},
  journal={Journal of Parallel and Distributed Computing}, volume={198}, pages={105027}, year={2025}}

@inproceedings{mcmahan2017,
  title={Communication-efficient learning of deep networks from decentralized data},
  author={McMahan, Brendan and Moore, Eider and others},
  booktitle={Proceedings of the 20th International Conference on Artificial Intelligence and Statistics}, pages={1273--1282}, year={2017}}

@inproceedings{fang2020,
  title={Local model poisoning attacks to {Byzantine}-robust federated learning},
  author={Fang, Minghong and Cao, Xiaoyu and others},
  booktitle={Proceedings of the 29th USENIX Security Symposium}, pages={1605--1622}, year={2020}}

@article{pillutla2022,
  title={Robust aggregation for federated learning},
  author={Pillutla, Krishna and Kakade, Sham M. and Harchaoui, Zaid},
  journal={IEEE Transactions on Signal Processing}, volume={70}, pages={1142--1154}, year={2022},
  doi={10.1109/TSP.2022.3153135}}

@article{li2022smartgrid,
  title={Detection of false data injection attacks in smart grid: A secure federated deep learning approach},
  author={Li, Yang and Wei, Xinhao and others},
  journal={IEEE Transactions on Smart Grid}, volume={13}, number={6}, pages={4862--4872}, year={2022}}

@article{zhang2026smartgrid,
  title={Targeted adversarial poisoning attack against robust aggregation in federated learning for smart grids},
  author={Zhang, Guihai and Sikdar, Biplab},
  journal={IEEE Transactions on Smart Grid}, volume={17}, number={2}, pages={1519--1534}, year={2026},
  doi={10.1109/TSG.2025.3629243}}

@inproceedings{li2019rsa,
  title={{RSA}: {Byzantine}-robust stochastic aggregation methods for distributed learning from heterogeneous datasets},
  author={Li, Liping and Xu, Wei and others},
  booktitle={Proceedings of the AAAI Conference on Artificial Intelligence}, number={01}, pages={1544--1551}, year={2019}}

@article{hoeffding1963,
  title={Probability inequalities for sums of bounded random variables},
  author={Hoeffding, Wassily},
  journal={Journal of the American Statistical Association}, volume={58}, number={301}, pages={13--30}, year={1963},
  doi={10.1080/01621459.1963.10500830}}
